\pdfoutput=1

\documentclass[11pt]{article}

\usepackage{acl}

\usepackage{times}
\usepackage{latexsym}

\usepackage[T1]{fontenc}

\usepackage[utf8]{inputenc}

\usepackage{microtype}

\usepackage{inconsolata}

\usepackage{graphicx}

\usepackage{amsmath}
\usepackage{amssymb}
\usepackage{xspace}
\usepackage{enumitem}

\tcbuselibrary{breakable}

\newcommand{\ourmethod}[0]{GraphReader\xspace}

\newcommand{\ie}{\emph{i.e.,}\xspace}
\newcommand{\aka}{\emph{i.e.,}\xspace}

\newcommand{\ignore}[1]{}

\title{
GraphReader: Building Graph-based Agent to Enhance \\Long-Context Abilities of Large Language Models
}

\author{
Shilong Li$^{*1}$,
Yancheng He$^{*1}$,
Hangyu Guo$^{*1}$,
Xingyuan Bu$^{*\dag \ddag 1}$,
Ge Bai$^{1}$,
Jie Liu$^{2,3}$,
\\
{\bf Jiaheng Liu$^{1}$, Xingwei Qu$^{4}$, Yangguang Li$^{3}$, Wanli Ouyang$^{2,3}$, Wenbo Su$^{1}$, Bo Zheng$^{1}$} \\
$^1$Alibaba Group\ \ \ 
$^2$The Chinese University of Hong Kong\\
$^3$Shanghai AI Laboratory\ \ \
$^4$University of Manchester\\
{\tt zhuli.lsl@taobao.com, xingyuanbu@gmail.com}
}

\begin{document}
\maketitle
\let\thefootnote\relax\footnotetext{$*$ First four authors contributed equally.}
\let\thefootnote\relax\footnotetext{$\dag$ Corresponding Author. $\ddag$  Project Leader.}
\begin{abstract}

Long-context capabilities are essential for large language models (LLMs) to tackle complex and long-input tasks. Despite numerous efforts made to optimize LLMs for long contexts, challenges persist in robustly processing long inputs.
In this paper, we introduce \ourmethod, a graph-based agent system designed to handle long texts by structuring them into a graph and employing an agent to explore this graph autonomously. Upon receiving a question, the agent first undertakes a step-by-step analysis and devises a rational plan. It then invokes a set of predefined functions to read node content and neighbors, facilitating a coarse-to-fine exploration of the graph. Throughout the exploration, the agent continuously records new insights and reflects on current circumstances to optimize the process until it has gathered sufficient information to generate an answer.
Experimental results on the LV-Eval dataset reveal that \ourmethod, using a 4k context window, consistently outperforms GPT-4-128k across context lengths from 16k to 256k by a large margin. Additionally, our approach demonstrates superior performance on four challenging single-hop and multi-hop benchmarks.
\end{abstract}

\section{Introduction}

\begin{figure}[t]
    \centering
    \includegraphics[width=0.99\linewidth]{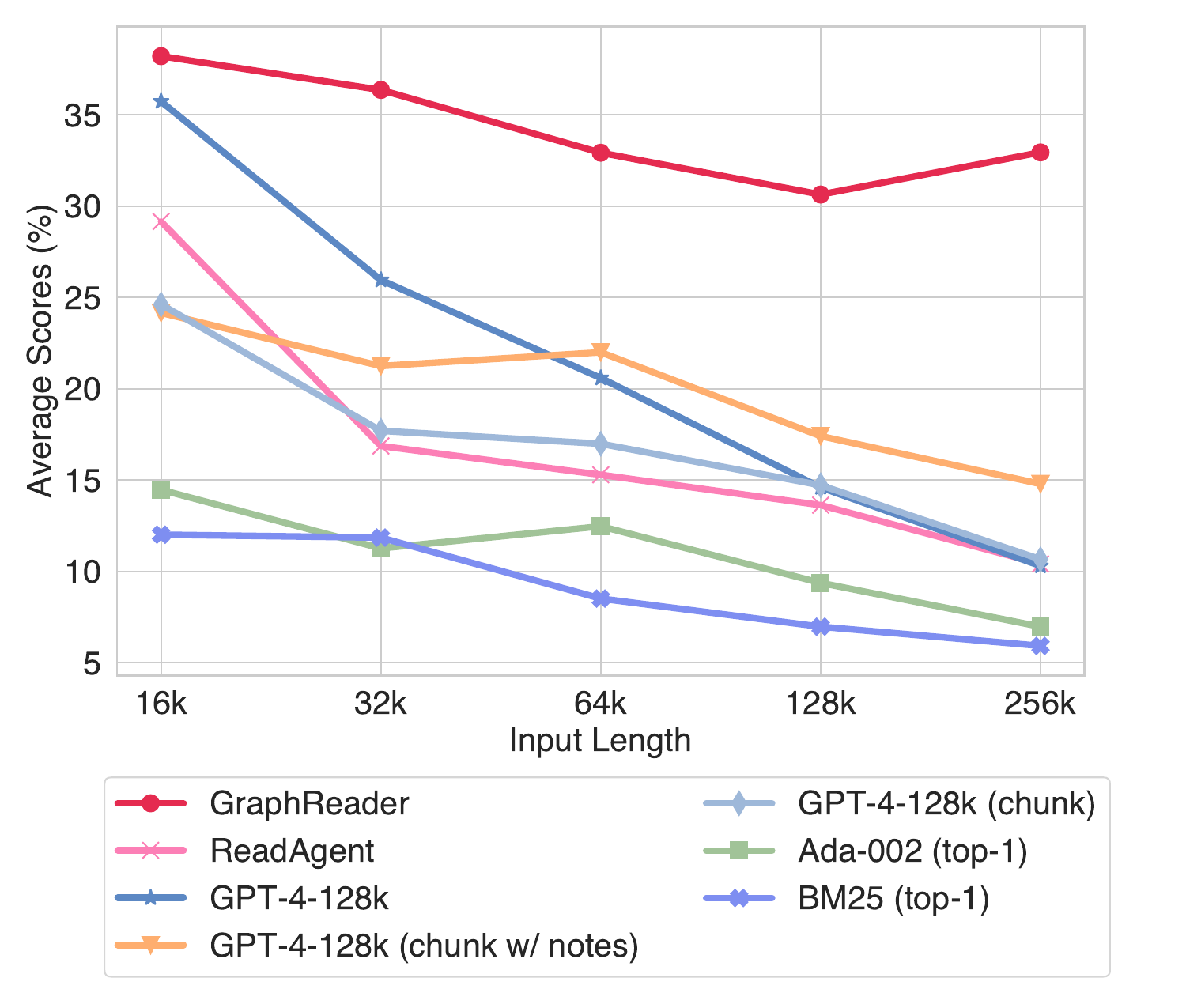}
    \caption{Performance on LV-Eval at 5 context length levels. GraphReader outperforms existing open-sourced and closed-source models while demonstrating a scalable performance in very long contexts. In contrast, other models exhibit a significant decrease in performance as context length increases.}
    \label{fig: motivation}
    \vspace{-5mm}
\end{figure}

Large language models~(LLMs) have made great progress on natural language understanding and generation~\citep{Zhao-2023-arxiv-survey, liu2024iterative, feng2022beyond, peng2020large, xv2022visual, peng2023gaia, bu2021gaia}. However, transformer-based LLMs still struggle in handling long contexts due to the limitation of context window and memory usage.

Current techniques for solving the long-context tasks of LLMs can be divided into two perspectives: 1) Model-level, which includes finetuning with modified positional embeddings~\cite{chen2023extending,zhu2023pose,peng2023yarn,ding2024longrope}, and applying transformer variants with modified attention mechanisms~\citep{dai2019transformer,munkhdalai2024leave,gu2023mamba}; 2) Agent-level, \aka employing retrieval-augmented LLM or agent to process long contexts with a limited context window LLM~\cite{nakano2021webgpt,lee2024human}.

However, model-level methods typically train LLMs with target length texts, posing challenges in constructing training datasets and incurring high training costs~\cite{zhu2023pose}. Additionally, long-context LLMs optimized with these methods tend to overlook crucial details in long contexts, known as ``lost in the middle''~\cite{liu2024lost}, limiting their ability to address complex tasks, such as multi-hop questions. Agent-level approaches transform input text into a tree~\citep{chen2023walking} or paginated pages~\citep{lee2024human}, failing to capture multi-hop and long-range dependencies, thus limiting their effectiveness on very long contexts, as shown in Figure~\ref{fig: motivation}.

To address these issues, we propose a graph-based agent named \textbf{\ourmethod}. As illustrated in Figure~\ref{fig: ourmethod}, \ourmethod first segments long texts into discrete chunks, extracts essential information, and compresses these into key elements and atomic facts. These key elements and facts are then used to construct a graph with nodes representing key elements and their associated atomic facts. This graph structure effectively captures long-range dependencies and multi-hop relationships within long text.
Subsequently, \ourmethod autonomously explores this graph using predefined functions, guided by a step-by-step rational plan. Based on a given question, the agent progressively accesses information from coarse key elements and atomic facts to detailed original text chunks, taking notes and reflecting until it gathers sufficient information to generate an answer.
In summary, our main contributions are threefold:

\begin{itemize}[leftmargin=4mm]
\vspace{-0.1cm}
\item We introduce \ourmethod, a novel agent system designed to organize long texts into a graph structure, leveraging predefined functions and notebook to facilitate planning and reflection during exploration.
\vspace{-0.1cm}

\item \ourmethod establishes a scalable long-context capability based on a 4k context window, demonstrating performance that is comparable to or surpasses GPT-4 with a 128k context window across varying context lengths.
\vspace{-0.1cm}

\item Extensive experiments conducted on four challenging benchmarks demonstrate that \ourmethod achieves superior performance in complex single-hop and multi-hop QA tasks.
\end{itemize}

\vspace{-0.3cm}
\section{Related Work}
\vspace{-0.3cm}
\paragraph{Long-Context LLMs}

Recent efforts~\citep{chen2023extending,ding2024longrope,peng2023yarn} have focused on positional interpolation (PI) to enhance long-context capabilities. However, these methods require training on full-length texts, leading to significant increases in data and training costs~\cite{chen2023longlora,fu2024data,bai2024longalign}. Thus, PoSE~\cite{zhu2023pose} and SkipAlign~\cite{wu2024long} investigate data skip strategy, but tend to neglect detailed information in long texts~\cite{liu2024lost,bai2024mt,wu2024conceptmath}. Furthermore, despite how extensively the context window is expanded, it remains constrained by a predefined fixed length. To address these limitations, transformer variants with modified attention mechanisms have been proposed \citep{dai2019transformer,gu2023mamba,munkhdalai2024leave}. However, these models are prone to losing earlier information.

\vspace{-0.1cm}
\paragraph{Retrieval}

Retrieval Augmented Generation (RAG) leverages an extensive database of documents to extract task-related information that aids in response generation. Many efforts investigate various levels of retrieval granularity, including tokens \citep{khandelwal2019generalization}, entities \citep{fevry2020entities,de2021mention}, and chunks \citep{LlamaIndex, LangChain}. Other approaches have explored diverse retrieval methods, such as BM25 \citep{rasooli-tetrault-2015} and learning-based strategies \citep{khattab2020colbert,sachan2023questions,sun2021long}. Despite its capabilities, RAG faces challenges in addressing complex questions due to difficulties in developing robust decision-making mechanisms. In contrast, we employ agents that use planning and reflection to gather essential information, effectively tackling complex problems.

\vspace{-0.1cm}
\paragraph{Agent for Retrieval}

Recent work has increasingly leveraged LLMs as agents to tackle complex problems, utilizing their strong planning and reflection abilities \cite{yao2022react,park2023generative}. These abilities have been applied to complex tasks such as function call \cite{li2023chain} and KGQA \cite{sun2023think, luo2023reasoning}. Agents are also capable of retrieving unstructured information. For example, WebGPT \cite{nakano2021webgpt} simulates human actions to search on internet for specific answers. Additionally, MemWalker \cite{chen2023walking} and PEARL \cite{sarthi2024raptor} organize documents into a tree structure, while ReadAgent \cite{lee2024human} condenses documents into a gist memory directory. However, these approaches often struggle with multi-hop questions. KGP \cite{wang2024knowledge} organizes documents into graphs, but it primarily uses the agent to generate queries, thereby not fully exploiting the agent's capabilities for planning and reflection.
\vspace{-0.2cm}
\section{Approach}
\vspace{-0.2cm}

\begin{figure*}[t]
    \centering
    \includegraphics[width=0.98\linewidth]{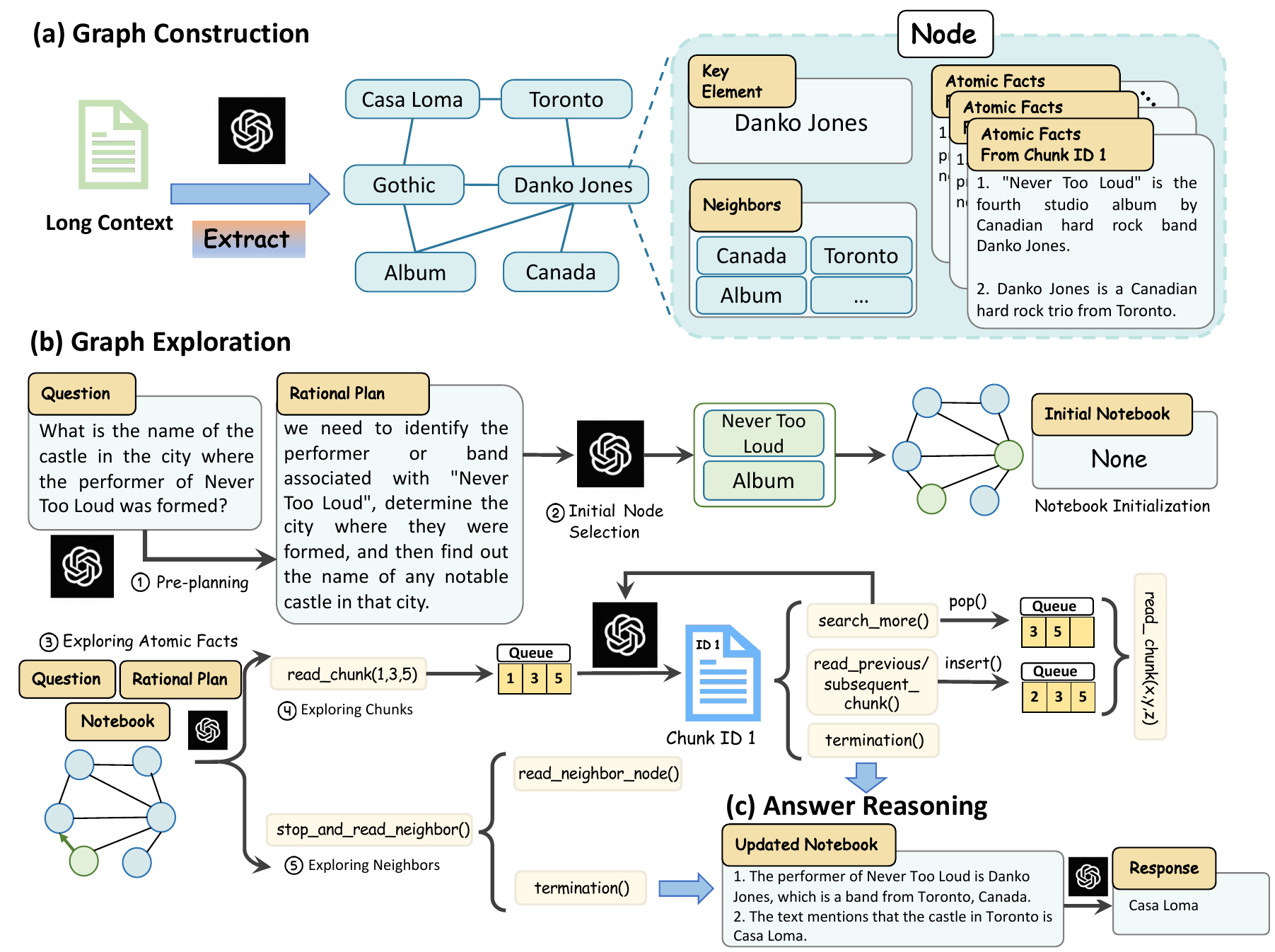}
    \caption{The illustration of our \ourmethod approach, consisting of graph construction, graph exploration, and answer reasoning.}
    \label{fig: ourmethod}
    \vspace{-4mm}
\end{figure*}

\subsection{Preliminary}

\ourmethod is built on a graph $\mathcal{G}=\left\{\mathcal{V}, \mathcal{E}\right\}$, where each node $v_i \in \mathcal{V}$ contains a key element $k_i$ and a set of summarized content, namely atomic facts $\mathcal{A}_i$. In other words, $v_i = \left\{k_i, \mathcal{A}_i\right\}$. And each edge $e_{ij} \in \mathcal{E}$ represents the relationship between nodes $v_i$ and $v_j$. This graph structure enables \ourmethod to capture global information from the input document $D$ within a limited context window, allowing it to decide whether to explore the current node in detail or jump to a neighboring node. During graph exploration, \ourmethod collects supporting facts and terminates the exploration once sufficient information has been gathered to answer the question. As illustrated in Figure~\ref{fig: ourmethod}, the entire process of \ourmethod consists of the following three phases: graph construction, graph exploration, and answer reasoning.
The prompts utilized in these three stages are detailed in Appendix~\ref{app: graphreader_prompt}, and a detailed example of our process can be found in Appendix \ref{app: case}.

\subsection{Graph Construction}

To extract nodes from a document \( D \) within the LLM's context limit, we first split \( D \) into chunks of maximum length \( L \) while preserving paragraph structure. For each chunk, we prompt the LLM to summarize it into atomic facts, the smallest indivisible facts that simplify the original text. We also prompt the LLM to extract key elements from each atomic fact like essential nouns, verbs, and adjectives. After processing all chunks, we normalize the key elements as described by \citet{lu2023instag} to handle lexical noise and granularity issues, creating a final set of key elements. We then construct each node \( v_i = \left( k_i, \mathcal{A}_i \right) \), where \( k_i \) is a key element and \( \mathcal{A}_i \) is the set of atomic facts corresponding to \( k_i \). Finally, we link two nodes \( v_i \) and \( v_j \) if key element \( k_i \) appears in \( \mathcal{A}_j \) and vice versa.

\subsection{Graph Exploration} 

\subsubsection{Agent Initialization}

Given a graph \(\mathcal{G}\) and a question \(Q\), our goal is to design an agent that can autonomously explore the graph using predefined functions. The agent begins by maintaining a notebook to record supporting facts, which are eventually used to derive the final answer. Then the agent performs two key initializations: defining the rational plan and selecting the initial node.

\paragraph{Rational Plan} To tackle complex real-world multi-hop questions, pre-planning the solution is crucial. The agent breaks down the original question step-by-step, identifies the key information needed, and forms a rational plan.

\paragraph{Initial Node} Choosing strategic starting points is essential for improving search efficiency. The agent evaluates the key elements of all nodes \(\mathcal{V}\) and selects \(N\) initial nodes based on the question and the rational plan.

\subsubsection{Exploration}

After selecting $N$ initial nodes as starting points, an agent explores each initial node by first exploring atomic facts, then chunks of the node. Next, it explores neighboring nodes, guided by the question and rational plan. The agent continuously updates the notebook with relevant information during the exploration process.

\paragraph{Exploring Atomic Facts}
It is impractical to include all original text chunks related to a node within the context window. Therefore, the agent employs a coarse-to-fine strategy, progressing from reading atomic facts to the original text, as all atomic facts can fit within the context window. Initially, all atomic facts associated with a node are grouped by their corresponding chunks, labeled with the respective chunk IDs, and fed to the agent. This allows the agent to capture an overview of each chunk by reading all groups of atomic facts. Meanwhile, the agent utilizes the question, rational plan, and notes in its notebook to reflect on the required clues and determine which chunk is likely to contain useful information. 
Subsequently, the agent is provided with two functions: 1) \textit{read\_chunk}, if the agent identifies certain chunks as valuable for further reading, it will complete the function parameters with the chunk IDs, \ie \textit{read\_chunk(List[ID])}, and append these IDs to a chunk queue. 2) \textit{stop\_and\_read\_neighbor}, conversely, if the agent deems that none of the chunks are worth further reading, it will finish reading this node and proceed to explore neighboring nodes.

\paragraph{Exploring Chunks}
When the chunk queue is non-empty, it indicates that the agent has identified multiple text chunks of interest. We then traverse the queue, reading each chunk. This step is essential because atomic facts merely summarize key information and provide brief clues, whereas specific details are best obtained directly from the original text chunks. 
While reading the chunks, the agent will once again consider the question and the plan, thinking about what can be added to the current notebook. Any supporting facts discovered will be recorded in the notebook. Depending on the updated notebook, the agent will then select one of the following four functions: 1) \textit{search\_more}, if supporting fact is insufficient, the agent will continue exploring chunks in the queue; 2) \textit{read\_previous\_chunk} and 3)\textit{read\_subsequent\_chunk}, due to truncation issues, adjacent chunks might contain relevant and useful information, the agent may insert these IDs to the queue; 4) \textit{termination}, if sufficient information has been gathered for answering the question, the agent will finish exploration.

\paragraph{Exploring Neighbors}
Once the atomic facts and chunk queue of the current node have been fully processed, it indicates that this node has been thoroughly explored, and the agent needs to access the next node. Taking into account the question, rational plan, and the content of the notebook, the agent checks all neighboring nodes, \ie key elements, and performs one of two functions: 1) \textit{read\_neighbor\_node}, the agent selects a neighboring node that might be helpful in answering the question and re-enters the process of exploring atomic facts and chunks; 2) \textit{termination}, the agent determines that none of the neighboring nodes contain useful information, it finish the exploration.

\subsection{Answer Reasoning}  
After $N$ agents have independently gathered information and stopped their exploration, we will compile all notes from each agent for reasoning and generating the final answer. Employing Chain-of-Thought~\cite{wei2022chain}, the LLM first analyzes each note by considering complementary information from other memories and using a majority voting strategy to resolve any inconsistencies. Ultimately, the LLM will consider all the available information to generate the final answer.

\section{Experiments}

\subsection{Experimental Settings}

\paragraph{Evaluation Benchmarks} We conduct experiments on two types of long-context QA benchmarks, including multi-hop long-context QA, \ie HotpotQA~\citep{yang-2018-emnlp-hotpotqa}, 2WikiMultihopQA~\citep{ho-2020-coling-2wikimultihopQA}, MuSiQue~\citep{Trivedi-2022-acltrans-musique}, and a single-hop long-context QA benchmark, \ie NarrativeQA~\citep{Kocisky-2018-acltrans-narrativeqa} from LongBench~\cite{Bai2023LongBenchAB}. Additionally, we also incorporate HotpotWikiQA-mixup from LV-Eval~\citep{Yuan2024LVEvalAB}, a multi-hop benchmark that features five levels of text length: 16k, 32k, 64k, 128k, and 256k. Table~\ref{tab: datasets} presents the statistics about these benchmarks, and detailed information is provided in Appendix~\ref{app: datasets}.

\paragraph{Evaluation Metrics} We employ several automatic evaluation metrics, \ie $F_1$ score, Exact Match (EM) score, and an optimized $F_1$* score, as introduced by LV-Eval~\cite{Yuan2024LVEvalAB}. Specifically, $F_1$* first computes the recall of golden answer keywords and only calculates the $F_1$ score if it exceeds a certain threshold. Otherwise, the score defaults to zero. Despite the cost-effectiveness of automatic metrics, their accuracy may be affected by the response format. Hence, we implement LLM Raters for answer correctness evaluation using an LLM, denoted as LLM-Rating-1 (LR-1) and LLM-Rating-1 (LR-2), following ReadAgent~\citep{lee2024human}. Details on the evaluation metrics can be found in Appendix \ref{app: eval_details}.

\paragraph{Baseline Methods} We compare our approach with the following baselines: retrieval augmented generation (RAG), long-context LLM, and agent-based methods.
(1) \textbf{RAG}: We choose Okapi BM25~\citep{Robertson2009ThePR} or OpenAI API embedding model Ada-002~\footnote{https://platform.openai.com/docs/guides/embeddings/\linebreak embedding-models} to retrieve the chunks most relevant to the question and employ GPT-4-128k (\texttt{gpt-4-1106-preview}) to read retrieved chunks and answer the question. In addition to traditional RAG methods, we also compared GraphRAG~\citep{graphrag} and LongRAG~\citep{Jiang2024LongRAGER}, which utilize LLM to enhance RAG ability.
(2) \textbf{Long-context LLM}: We select GPT-4-128k for directly reading full text when the text content fits within the input window, or for segmenting the text into chunks for sequential reading.
(3) \textbf{Agent-based Method}: We select ReadAgent~\citep{lee2024human} and PEARL~\citep{sun-etal-2024-pearl}, which employ an agent-based system for the execution of retrieval and reading processes for long-context QA.
The detailed description of these methods is provided in Appendix~\ref{app: baseline}.

\paragraph{Implementation Details}

In our experiments, we employ GPT-4-128k for both our method and baseline approaches, setting the temperature to 0.2. 
For \ourmethod, the input window size is configured to 4k tokens unless stated otherwise. We limit the maximum chunk size to 2k tokens, initiate searches from 5 initial nodes, and impose a function call limit of 10 for each search path.

\begin{table}[!t]
\centering
\resizebox{\linewidth}{!}{
\begin{tabular}{llrrc}
\bottomrule
                  \textbf{Task} & \textbf{Dataset}  & Avg \#Tokens & Max \#Tokens & \#Samples  \\ \midrule
\multirow{4}{*}{\textbf{Multi-hop QA}} & HotpotQA  & 9.4k & 15.9k & 300  \\
                     & 2WikiMultihopQA  & 8.8k & 15.9k & 300 \\
                    & MuSiQue  & 15.5k & 16.0k & 200 \\
                    & HotpotWikiQA-mixup  & 142.4k & 370.8k & 250 \\ \midrule
\textbf{Single-hop QA} & NarrativeQA  & 29.7k & 63.7k & 200 \\\bottomrule
\end{tabular}
}
\caption{
The statistics of benchmarks employed in our evaluation. The token number is calculated using the GPT-4 tokenizer from the TikToken\protect\footnotemark. \#Samples denote the total number of benchmarks.
}
\label{tab: datasets}
\end{table}
\footnotetext{https://github.com/openai/tiktoken}

\begin{table*}[!t]
\renewcommand{\arraystretch}{1.1}
\centering
\resizebox{\textwidth}{!}{
    \begin{tabular}{lcrrrr|rrrr|rrrr|rrrr}
    \toprule
    \multirow{2}{*}{\textbf{Method}} & \textbf{Input} & \multicolumn{4}{c}{\textbf{HotpotQA}} & \multicolumn{4}{c}{\textbf{2WikiMultihopQA}} & \multicolumn{4}{c}{\textbf{MuSiQue}} &  \multicolumn{4}{c}{\textbf{NarrativeQA}}\\ \cline{3-18}
     & \textbf{Window} & LR-1 & LR-2 & EM & $F_1$ & LR-1 & LR-2 & EM & $F_1$ & LR-1 & LR-2 & EM & $F_1$ & LR-1 & LR-2 & EM & $F_1$\\ \midrule
    BM25 (top-1) & $4k$ & 57.7 & 63.0 & 33.7 & 43.8 & 36.0 & 39.0 & 25.0 & 30.4 & 33.0 & 36.5 & 19.0 & 23.9 & 29.5 & 34.5 & 4.0 & 11.3 \\
    BM25 (top-3) & $4k$ & 74.7 & 78.3 & 45.7 & 58.5 & 59.7 & 62.0 & 42.3 & 51.9 & 43.5 & 49.5 & 25.0 & 31.1 & 44.5 & 52.5 & 7.0 & 20.5 \\ \midrule
    Ada-002 (top-1) & $4k$ & 63.0 & 70.7 & 40.0 & 53.2 & 57.0 & 59.3 & 41.0 & 49.4 & 34.5 & 37.0 & 20.0 & 26.6 & 37.5 & 46.5 & 5.0 & 15.5 \\
    Ada-002 (top-3) & $4k$ & 72.0 & 77.3 & 45.0 & 58.1 & 65.7 & 66.7 & 44.7 & 55.3 & 40.0 & 45.5 & 24.5 & 32.1 & 45.5 & 53.0 & 7.5 & 19.5 \\ \midrule
    GPT-4-128k & $128k$ & \underline{83.3} & \underline{88.3} & \underline{53.0} & \underline{68.4} & 77.3 & 80.0 & \underline{58.7} & \underline{70.0} & 52.0 & 59.5 & 33.5 & 42.7 & \underline{63.5} & \underline{77.0} & 11.5 & \underline{29.4}  \\
    GPT-4-128k (chunk) & $4k$ & 71.3 & 74.7 & 45.7 & 59.5 & 59.3 & 62.3 & 40.7 & 50.5 & 41.0 & 43.0 & 23.0 & 32.1 & 58.0 & 69.5 & 9.50 & 25.5  \\
    GPT-4-128k (chunk w/ notes) & $4k$ & 72.3 & 76.7 & 45.7 & 59.5 & 65.7 & 68.7 & 46.3 & 56.6 & 39.5 & 43.0 & 25.0 & 32.5 & 56.5 & 65.0 & 8.5 & 24.3  \\\midrule
    ReadAgent & $128k$ & 72.3 & 78.7 & 48.0 & 62.0 & \underline{79.0} & \underline{81.0} & 52.7 & 63.7 & \underline{54.5} & \underline{61.0} & \underline{35.0} & \underline{45.1} & 63.0 & 75.5 & 5.0 & 18.9 \\
    Pearl & $128k$ & 74.7 & 79.0 & 46.3 & 60.4 & 70.0 & 71.0 & 46.0 & 57.6 & 45.0 & 51.5 & 23.0 & 33.3 & 43.5 & 48.0 & 7.5 & 16.2 \\
    LongRAG & $128k$ & 75.7 & 78.3 & 48.7 & 63.9 & 73.0 & 75.0 & 51.3 & 63.5 & 49.0 & 54.5 & 31.0 & 40.3 & 60.5 & 69.0 & 15.0 & 27.0 \\
    GraphRAG & $128k$ & 73.7 & 80.3 & 49.7 & 59.7 & 67.7 & 71.3 & 42.3 & 53.9 & 46.5 & 56.0 & 21.5 & 31.2 & 52.0 & 66.5 & \underline{15.0} & 23.1 \\
    \ourmethod & $4k$ &\textbf{84.3} & \textbf{89.7} & \textbf{55.0} & \textbf{70.0} & \textbf{83.7} & \textbf{87.0} & \textbf{59.3} & \textbf{70.1} & \textbf{59.0} & \textbf{63.5} & \textbf{38.0} & \textbf{47.4} & \textbf{65.0} & \textbf{80.0} & \textbf{15.5} & \textbf{29.8} \\
    \midrule
    Golden & $4k$ & 92.3 & 93.7 & 57.0 & 73.8 & 88.3 & 89.7 & 63.0 & 73.4 & 66.0 & 69.0 & 45.0 & 56.0 & - & - & - & - \\  \midrule
    \end{tabular}
}
\caption{Performance~(\%) comparison of different baselines on datasets from LongBench. The best performance and the second-best performance are denoted in bold and underlined fonts, respectively. ``Golden'' denotes the settings in which we add question and its supporting facts to LLM directly.}
\label{tab: main_result_1}
\end{table*}
\begin{table*}[!t]
\renewcommand{\arraystretch}{1.1}
\centering
\resizebox{\textwidth}{!}{
\begin{tabular}{lcrrr|rrr|rrr|rrr|rrr}
\midrule
 \multirow{3}{*}{\textbf{Method}} & \multirow{2}{*}{\textbf{Input}} & \multicolumn{15}{c}{\textbf{HotpotWikiQA-mixup}} \\\cline{3-17}
 & \multirow{2}{*}{\textbf{Window}} & \multicolumn{3}{c}{\textbf{16k}} & \multicolumn{3}{c}{\textbf{32k}} & \multicolumn{3}{c}{\textbf{64k}} & \multicolumn{3}{c}{\textbf{128k}} & \multicolumn{3}{c}{\textbf{256k}} \\ \cline{3-17}
& & LR-1 & LR-2 & $F_1$* & LR-1 & LR-2 & $F_1$* & LR-1 & LR-2 & $F_1$* & LR-1 & LR-2 & $F_1$* & LR-1 & LR-2 & $F_1$*\\ \midrule
    BM25 (top-1) & $4k$ & 10.0 & 16.0 & 12.0 & 16.0 & 18.0 & 11.9 & 6.0 & 8.0 & 8.5 & 10.0 & 8.0 & 7.0 & 14.0 & 20.0 & 5.9 \\
    BM25 (top-3) & $4k$ & 16.0 & 22.0 & 13.9 & 18.0 & 28.0 & 13.3 & 16.0 & 18.0 & 11.8 & 12.0 & 16.0 & 11.8 & 12.0 & 22.0 & 9.3 \\ \midrule
    Ada-002 (top-1) & $4k$ & 10.0 & 12.0 & 14.5 & 14.0 & 18.0 & 11.3 & 10.0 & 12.0 & 12.5 & 12.0 & 14.0 & 9.4 & 8.0 & 8.0 & 7.0 \\
    Ada-002 (top-3) & $4k$ & 24.0 & 28.0 & 21.3 & 20.0 & 30.0 & 19.8 & 14.0 & 20.0 & 12.9 & 16.0 & 20.0 & 12.0 & 14.0 & 18.0 & 10.8 \\  \midrule
    GPT-4-128k & $128k$ & {\underline {38.0}} & {\underline {38.0}} & {\underline {35.7}} & {\underline {26.0}} & {\underline {30.0}} & {\underline {26.0}} & 22.0 & 24.0 & 20.6 & 16.0 & 16.0 & 14.6 & 14.0 & 16.0 & 10.3 \\
    GPT-4-128k (chunk) & $4k$ & 18.0 & 22.0 & 24.6 & 16.0 & 20.0 & 17.7 & 20.0 & 24.0 & 17.0 & 20.0 & 24.0 & 14.7 & {\underline {28.0}} & {\underline {30.0}} & 10.7 \\
    GPT-4-128k (chunk w/ notes) & $4k$ & 22.0 & 32.0 & 24.2 & 26.0 & 30.0 & 21.3 & {\underline {28.0}} & {\underline {32.0}} & {\underline {22.0}} & {\underline {24.0}} & {\underline {26.0}} & {\underline {17.4}} & 26.0 & 26.0 & {\underline {14.8}} \\ \midrule
    ReadAgent & $128k$ & 24.0 & 26.0 & 29.2 & 20.0 & 22.0 & 16.9 & 24.0 & 30.0 & 15.3 & 14.0 & 18.0 & 13.6 & 20.0 & 22.0 & 10.4 \\
    \ourmethod & $4k$ & \textbf{42.0} & \textbf{42.0} & \textbf{38.2} & \textbf{32.0} & \textbf{38.0} & \textbf{36.4} & \textbf{30.0} & \textbf{36.0} & \textbf{32.9} & \textbf{28.0} & \textbf{34.0} & \textbf{30.6} & \textbf{30.0} & \textbf{38.0} & \textbf{33.0} \\
\midrule
\end{tabular}
}
\caption{Performance~(\%) of different baselines on datasets from LV-Eval, where $F_1$* donates LV-Eval's optimized $F_1$. The best performance and the second-best performance are denoted in bold and underlined fonts, respectively. We truncate to keep the longest possible initial fragment while preserving paragraph structure, in contexts that exceed the input window ($128k$ and $256k$) for GPT-4-128k.}
\label{tab: main_result_2}
\end{table*}
\subsection{Main Results}

\ignore{\paragraph{Evaluation on Single and Multi-hop Long-context Tasks}} The results of three types of methods on four multi-hop long-context benchmarks and one single-hop long-context benchmark are shown in Table~\ref{tab: main_result_1} and Table~\ref{tab: main_result_2}. Based on the results, we have the following findings:

\paragraph{Results of RAG methods}
As the results shown in Table~\ref{tab: main_result_1}, RAG methods based on BM25 and Ada-002 exhibit the worst performance in comparison to long-context LLM and agent-based methods. A possible reason is that text retrieval has difficulty recalling all chunks that contain the supporting facts for answering the input question. Although increasing the number of recalled chunks could improve the performance of text retrieval, the context window will limit the effectiveness of these RAG methods. 

\paragraph{Results of Long-Context LLMs}

From the results shown in Table~\ref{tab: main_result_1}, we can see that employing GPT-4-128k to directly answer the question with long contexts significantly outperforms RAG methods and even outperforms ReadAgent on three long-context benchmarks. This is because of the superior performance of GPT-4-128k in processing long texts and executing multi-hop reasoning tasks. Additionally, the lengths of these four benchmarks are significantly shorter than the 128k context window, thereby mitigating the impact of ``lost in the middle'' on the model's performance. 

\ignore{Moreover, we also observe that GPT-4, despite having a constrained 4k context window, can process input contexts that significantly exceed its context window limitations through sequential reading. By incorporating a memory module, sequential reading with LLM can significantly boost the performance on multi-hop benchmarks. The reason may be that carrying memory with historical information can facilitate LLM in aggregating previous supporting facts. Conversely, for the single-hop benchmark, NarrativeQA, integrating memory results in performance degradation. The reason may be that previous memory introduces noise to LLM and affects its responses. }

\paragraph{Results of Agent-based Methods}

By comparing our approach with all baselines in Table~\ref{tab: main_result_1}, it is obvious that our approach consistently performs better than them on four long-context benchmarks and demonstrates superior performance in multi-hop long-context tasks. In our approach, benefiting from the graph's ability to capture the relationships between detailed information, our method can identify crucial information and search for the supporting facts for the input question efficiently. This strategy significantly boosts the agent's capability in multi-hop reasoning and capturing long-range dependencies of key information in a long context. Moreover, the results in Table~\ref{tab: main_result_1} show that ReadAgent, with a 128k context window setup, underperforms \ourmethod with a 4k context window and even performs worse than GPT-4-128k full-text reading. 
We attribute this to ReadAgent's strategy of excessively compressing the original texts into gist memories, and feeding all mixed memories to the model for page number selection. Compared to our \ourmethod, the strategy of ReadAgent may restrict the agent's ability to identify specific details and capture intrinsic connections among key elements in a long context, consequently affecting its overall performance.
This further indicates that our approach can more efficiently unlock the capabilities of constrained context window LLMs in processing long context. Additionally, we observe that the performance of our method closely matches that achieved by directly supplying supporting facts to the LLM (\ie Golden in Table~\ref{tab: main_result_1}). This is because our method incorporates not only pre-planning, reflection, and various actions but also the usage of a graph containing key information, facilitating the agent to search for the correct supporting facts.

For additional results on benchmarks relevant to real-world scenarios, please refer to the appendix~\ref{app: addtional_exp}.

\paragraph{Evaluation on Extremely Long Context Tasks} As shown in previous experiments, it demonstrates the effectiveness of employing a limited context window LLM for long-context tasks with our \ourmethod. Here, we would like to study the impact of extremely long context on our \ourmethod. As shown in Table~\ref{tab: main_result_2}, compared with all baselines, our \ourmethod not only consistently outperforms these methods across text lengths ranging from 16k to 256k tokens but also exhibits robustness with the expansion of context length. It indicates that our method is still effective in handling extremely long texts by graph exploration with limited context window LLMs. With the increase in the length of the input context, the performance of GPT-4-128k full-text reading degrades gradually. As a comparison, our method achieves a performance gain of 10.53\% relatively on LR-1 over GPT-4-128k full-text reading under 16k context length. With the context length increasing to 128k, our method achieves a performance gain of 75.00\% relatively over GPT-4-128k. This can be attributed to the fact that as the context length increases, the impact of the ``lost in the middle'' effect on GPT-4-128k becomes progressively more severe. Secondly, we observe that ReadAgent significantly underperforms our method in handling extremely long contexts. This is because the lack of detailed information about the content of each page can make page selection very difficult for ReadAgent, especially when dealing with extremely long contexts. This further demonstrates that our method can effectively address the challenges of processing extremely long context with limited context window LLMs by exploring graphs containing fine-grained information.

\begin{table}[!t]
\centering
\resizebox{\columnwidth}{!}{
    \begin{tabular}{clccc}
    \bottomrule
    \multirow{2}{*}{\textbf{Dataset}} & \multirow{2}{*}{\textbf{Method}} & \multicolumn{3}{c}{\textbf{Results(\%)}}\\ \cline{3-5} 
     & & LR-1 & LR-2 & $F_1$ \\ \midrule
    \multirow{3}{*}{\textbf{HotpotQA}} & \ourmethod & \textbf{84.3} & \textbf{89.7} & \textbf{70.0} \\
        & \quad w/o Rational Plan & 81.7 & 87.7 & 63.8 \\
        & \quad w/o Node Selection & 66.0 & 71.7 & 54.1 \\\midrule
    \multirow{3}{*}{\textbf{2WikiMultihopQA}}& \ourmethod & \textbf{83.7} & \textbf{87.0} & \textbf{70.1} \\
        & \quad w/o Rational Plan & 81.3 & 86.0 & 65.4 \\
        & \quad w/o Node Selection & 65.3 & 68.7 & 49.7 \\\midrule
    \multirow{3}{*}{\textbf{MuSiQue}} & \ourmethod & \textbf{59.0} & \textbf{63.5} & \textbf{47.4} \\
    & \quad w/o Rational Plan & 56.0 & 61.0 & 42.4 \\
    & \quad w/o Node Selection & 35.0 & 38.5 & 25.2 \\\midrule
    \multirow{3}{*}{\textbf{NarrativeQA}} & \ourmethod & \textbf{65.0} & \textbf{80.0} & \textbf{29.8}  \\
        & \quad w/o Rational Plan & 63.0 & 78.5 & 26.6 \\
        & \quad w/o Node Selection & 53.0 & 65.5 & 24.0 \\
    \bottomrule
    \end{tabular}
}
\vspace{-0.2cm}
\caption{The results of our ablation study. ``w/o Rational Plan'' refers to removing the rational plan in the agent initialization stage, and ``w/o Node Selection'' denotes applying the random selection of initial nodes and neighbor nodes in graph exploration.}
\label{tab: ablation}
\end{table}
\subsection{Ablation study}

\paragraph{The Effect of Rational Plan}

In the graph exploration stage, we introduce a rational plan to help the agent analyze complex input questions step by step, guiding the agent in exploring the graph. To verify the effectiveness of the rational plan, we removed it during agent initialization and conducted experiments on four long-context QA benchmarks. Table~\ref{tab: ablation} shows that the rational plan is effective in guiding the agent in node selection and exploration on the graph.

\paragraph{The Effect of Node Selection}
We conduct randomly selecting initial nodes and neighbor nodes experiments to demonstrate the necessity of our system in selecting which nodes to visit based on reasoning about the required information. As shown in Table \ref{tab: ablation}, random selection results in a significant performance drop, with an average decline of 18\%. This demonstrates that \ourmethod carefully considers node selection, leading to more reasonable and effective exploration.

\begin{figure}[!t]
\centering
\includegraphics[width=1\linewidth]{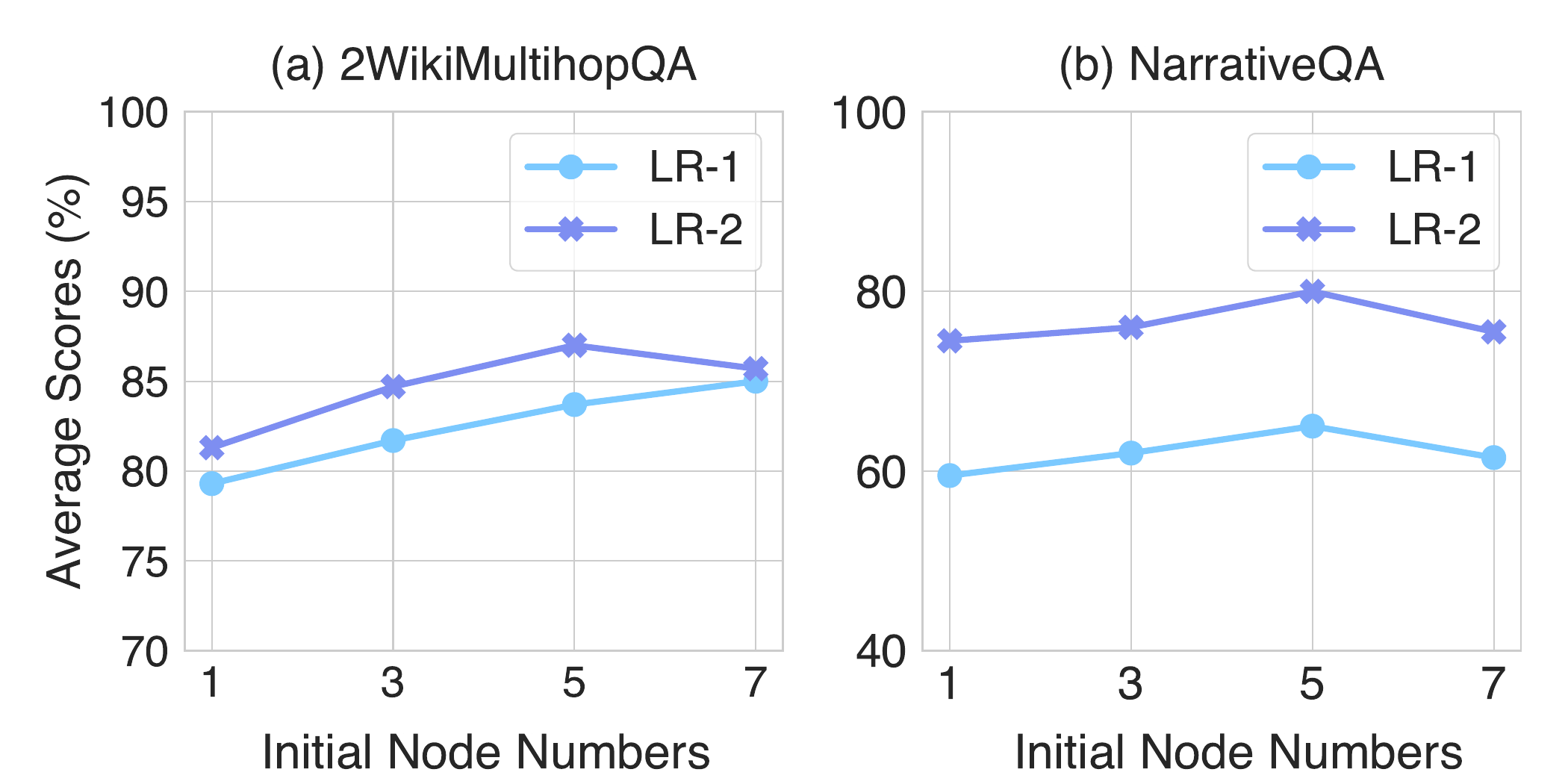}
\caption{Performance of \ourmethod with different initial node numbers on 2WikiMultihopQA and NarrativeQA. Results show the robustness of \ourmethod towards different initial node numbers.}
\label{fig: nodes}
\end{figure}

\paragraph{Impact of the Number of Initial Nodes}
We conduct experiments with different initial node counts on multi-hop and single-hop QA datasets to assess the effect of the number of initial nodes on \ourmethod's performance. The results are shown in Figure \ref{fig: nodes}. Increasing the number of nodes improves performance up to a certain point, with optimal performance at 5 initial nodes, which we set as the default. However, beyond this threshold, performance declines, especially in single-hop scenarios, likely due to increased noise from too many initial nodes.

\begin{figure}[!t]
\centering
\includegraphics[width=1.0\linewidth]{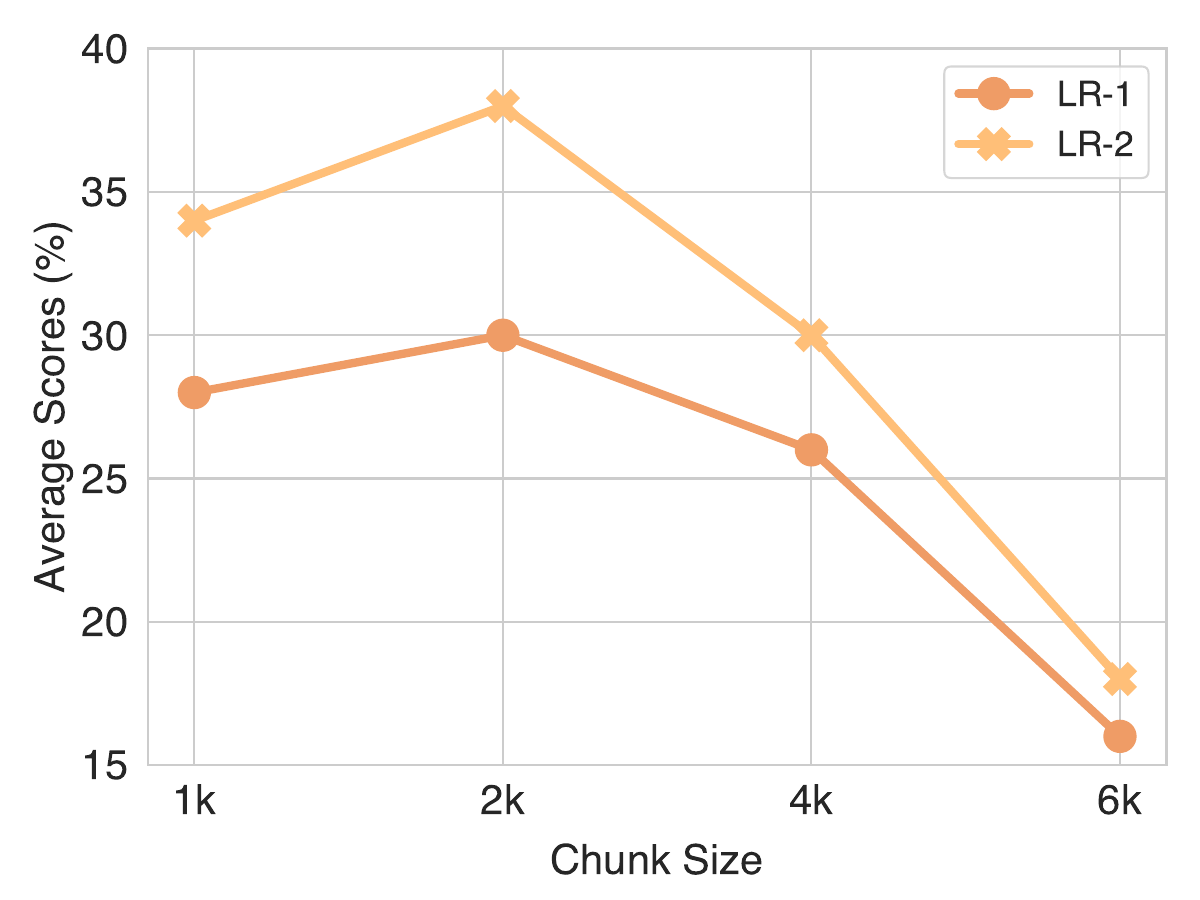}
\caption{The impact of chunk size $L$ of \ourmethod on the 256k length level of HotpotWikiQA-mixup.}
\label{fig: chunk_size_exp}
\end{figure}

\paragraph{Impact of the Chunk Size}
We investigate the impact of chunk size $L$ on \ourmethod's performance. As shown in Figure \ref{fig: chunk_size_exp}, the best performance is achieved with $L=2k$. When $L$ exceeds a certain threshold, performance declines because larger chunks cause the model to overlook essential details. Conversely, smaller chunks lead to more semantic truncation, hindering comprehension and accuracy in extracting atomic facts. Thus, we chose $L=2k$ as the default chunk size.

\subsection{Further Analysis}

\begin{table}[!t]
\resizebox{\columnwidth}{!}{
\begin{tabular}{ccc}
\toprule
    Method & Avg. Ctx. \#Tokens & Avg. Cost \#Tokens  \\\midrule
    ReadAgent & 358.3k & 48.7k  \\
    \ourmethod & 358.3k & 52.8k \\\bottomrule
    \end{tabular}
}
\caption{Comparison of token consumption per question between ReadAgent and \ourmethod on HotpotWikiQA-mixup-256k, where ``Avg. Ctx. \#Tokens'' refers to the average token number of the original dataset. The ``Avg. Cost \#Tokens'' comprise both input tokens and output tokens during exploration.}
\label{tab: consumption}
\end{table}

\paragraph{Cost Analysis}
To assess the inference cost of our approach, we compare the average token consumption of ReadAgent and \ourmethod for individual questions. As shown in Table \ref{tab: consumption}, \ourmethod uses only 1.08 times more tokens than ReadAgent (52.8k / 48.7k), yet achieves more than double the performance improvement, demonstrating its superiority. More importantly, our method has significant advantages in single-document multiple-query scenarios, where only one graph needs to be constructed. Subsequent QA can be performed on this graph, thereby reducing the overall token consumption.
\begin{figure}[!t]
\centering
\includegraphics[width=1\linewidth]{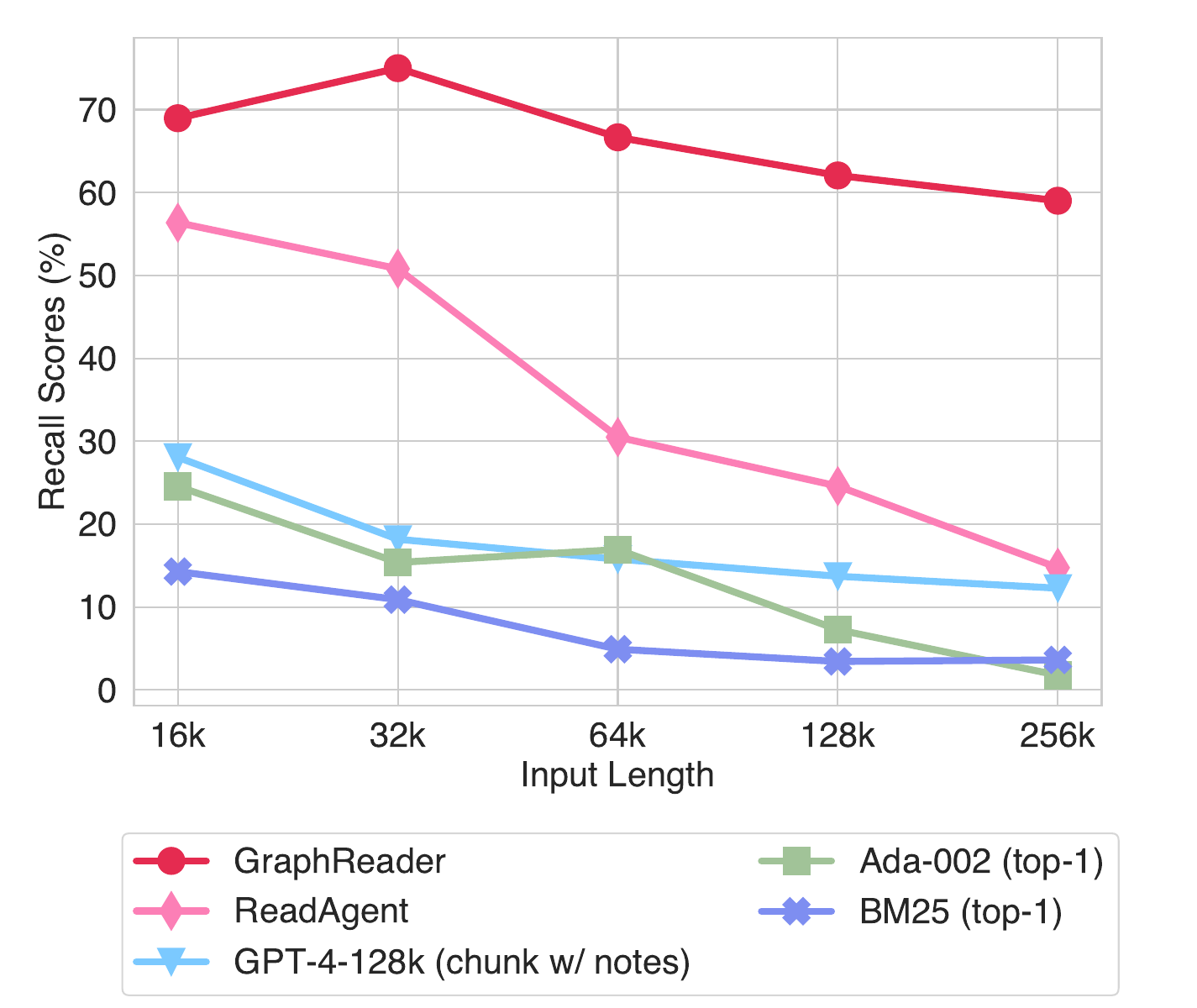}
\caption{Recall of supporting facts by different methods on HotpotWikiQA-mixup.}
\label{fig: recall}
\end{figure}

\paragraph{Recall Rate Analysis}
To evaluate our method's advantages in key information recall, we utilize GPT-4 to assess the recall of supporting facts on the HotpotWikiQA-mixup dataset. As shown in Figure~\ref{fig: recall}, our model consistently outperforms other baseline methods, regardless of the input length. As context length increases from 16k to 256k, recall of supporting facts declines across all methods. However, \ourmethod maintains around 60\% recall at 256k context length, in contrast to the significant degradation in ReadAgent. This demonstrates \ourmethod's scalability and effectiveness in processing long contexts.
Further details and evaluation prompts can be found in Appendix~\ref{app: recall}. 

To further demonstrate the recall rate of \ourmethod at different granularities, we calculate the recall rate of \emph{Supporting Facts} and \emph{Sample} granularity respectively using the same method, detailed in the Appendix~\ref{app: recall}. The granularity of supporting facts refers to the recall rate of all supporting facts across the entire dataset. As for sample granularity, a sample is considered to be recalled only if all of its supporting facts are recalled.
As shown in the Tabel~\ref{tab: recall}, the recall for the final notebook is slightly higher than the recall of atomic facts, which indicates that our method is capable of extracting more valid information from chunks during the exploration, indirectly reflecting its intelligence and effectiveness in exploration.

\begin{table}[!t]
\centering
\resizebox{0.75\columnwidth}{!}{
    \begin{tabular}{ccc}
    \bottomrule
    \multirow{2}{*}{\textbf{Source}}  & \multicolumn{2}{c}{\textbf{Recall(\%)}} \\ \cline{2-3} 
     &  SF-wise & Sample-wise \\ \midrule
    Atomic Facts & 76.4 & 64.7 \\
    Final Notebook & 90.5 & 85.3  \\
    \bottomrule
    \end{tabular}
}
\caption{\ourmethod's recall performance at different granularities on HotpotQA. ``SF-wise'' refers to the granularity of supporting facts, and ``Sample-wise'' refers to the granularity of sample evaluation.}
\label{tab: recall}
\end{table}

\section{Conclusion}
This paper introduces \ourmethod, a graph-based agent designed to enhance the long-context capabilities of large language models. \ourmethod organizes long texts into graph structures and employs an autonomous agent to explore the graph, successfully establishing long-range dependencies within a relatively small 4k context window. Experiments demonstrate that \ourmethod outperforms GPT-4 with a 128k input length across various long-context single-hop and multi-hop question-answering benchmarks.

\section{Limitations}
Firstly, \ourmethod is constructed using an off-the-shelf GPT-4 API. Since it is close-sourced, there may be potential restrictions such as limits on Queries Per Second (QPS) and regional constraints. Therefore, future work will involve collecting data, training models, and making them open-source to contribute to the wider community. Secondly, the efficiency of the agent depends on its planning and reasoning capabilities. Future research will also explore enhancements of these features to improve the effectiveness of our method.

\bibliography{custom}

\begin{thebibliography}{54}
\providecommand{\natexlab}[1]{#1}

\bibitem[{Bai et~al.(2024{\natexlab{a}})Bai, Liu, Bu, He, Liu, Zhou, Lin, Su, Ge, Zheng et~al.}]{bai2024mt}
Ge~Bai, Jie Liu, Xingyuan Bu, Yancheng He, Jiaheng Liu, Zhanhui Zhou, Zhuoran Lin, Wenbo Su, Tiezheng Ge, Bo~Zheng, et~al. 2024{\natexlab{a}}.
\newblock Mt-bench-101: A fine-grained benchmark for evaluating large language models in multi-turn dialogues.
\newblock \emph{arXiv preprint arXiv:2402.14762}.

\bibitem[{Bai et~al.(2024{\natexlab{b}})Bai, Lv, Zhang, He, Qi, Hou, Tang, Dong, and Li}]{bai2024longalign}
Yushi Bai, Xin Lv, Jiajie Zhang, Yuze He, Ji~Qi, Lei Hou, Jie Tang, Yuxiao Dong, and Juanzi Li. 2024{\natexlab{b}}.
\newblock Longalign: A recipe for long context alignment of large language models.
\newblock \emph{arXiv preprint arXiv:2401.18058}.

\bibitem[{Bai et~al.(2023)Bai, Lv, Zhang, Lyu, Tang, Huang, Du, Liu, Zeng, Hou, Dong, Tang, and Li}]{Bai2023LongBenchAB}
Yushi Bai, Xin Lv, Jiajie Zhang, Hong Lyu, Jiankai Tang, Zhidian Huang, Zhengxiao Du, Xiao Liu, Aohan Zeng, Lei Hou, Yuxiao Dong, Jie Tang, and Juanzi Li. 2023.
\newblock \href {https://api.semanticscholar.org/CorpusID:261245264} {Longbench: A bilingual, multitask benchmark for long context understanding}.
\newblock \emph{ArXiv}, abs/2308.14508.

\bibitem[{Bu et~al.(2021)Bu, Peng, Yan, Tan, and Zhang}]{bu2021gaia}
Xingyuan Bu, Junran Peng, Junjie Yan, Tieniu Tan, and Zhaoxiang Zhang. 2021.
\newblock Gaia: A transfer learning system of object detection that fits your needs.
\newblock In \emph{Proceedings of the IEEE/CVF Conference on Computer Vision and Pattern Recognition}, pages 274--283.

\bibitem[{Chen et~al.(2023{\natexlab{a}})Chen, Pasunuru, Weston, and Celikyilmaz}]{chen2023walking}
Howard Chen, Ramakanth Pasunuru, Jason Weston, and Asli Celikyilmaz. 2023{\natexlab{a}}.
\newblock Walking down the memory maze: Beyond context limit through interactive reading.
\newblock \emph{arXiv preprint arXiv:2310.05029}.

\bibitem[{Chen et~al.(2023{\natexlab{b}})Chen, Wong, Chen, and Tian}]{chen2023extending}
Shouyuan Chen, Sherman Wong, Liangjian Chen, and Yuandong Tian. 2023{\natexlab{b}}.
\newblock Extending context window of large language models via positional interpolation.
\newblock \emph{arXiv preprint arXiv:2306.15595}.

\bibitem[{Chen et~al.(2023{\natexlab{c}})Chen, Qian, Tang, Lai, Liu, Han, and Jia}]{chen2023longlora}
Yukang Chen, Shengju Qian, Haotian Tang, Xin Lai, Zhijian Liu, Song Han, and Jiaya Jia. 2023{\natexlab{c}}.
\newblock Longlora: Efficient fine-tuning of long-context large language models.
\newblock \emph{arXiv preprint arXiv:2309.12307}.

\bibitem[{Dai et~al.(2019)Dai, Yang, Yang, Carbonell, Le, and Salakhutdinov}]{dai2019transformer}
Zihang Dai, Zhilin Yang, Yiming Yang, Jaime Carbonell, Quoc~V Le, and Ruslan Salakhutdinov. 2019.
\newblock Transformer-xl: Attentive language models beyond a fixed-length context.
\newblock \emph{arXiv preprint arXiv:1901.02860}.

\bibitem[{De~Jong et~al.(2021)De~Jong, Zemlyanskiy, FitzGerald, Sha, and Cohen}]{de2021mention}
Michiel De~Jong, Yury Zemlyanskiy, Nicholas FitzGerald, Fei Sha, and William Cohen. 2021.
\newblock Mention memory: incorporating textual knowledge into transformers through entity mention attention.
\newblock \emph{arXiv preprint arXiv:2110.06176}.

\bibitem[{Ding et~al.(2024)Ding, Zhang, Zhang, Xu, Shang, Xu, Yang, and Yang}]{ding2024longrope}
Yiran Ding, Li~Lyna Zhang, Chengruidong Zhang, Yuanyuan Xu, Ning Shang, Jiahang Xu, Fan Yang, and Mao Yang. 2024.
\newblock Longrope: Extending llm context window beyond 2 million tokens.
\newblock \emph{arXiv preprint arXiv:2402.13753}.

\bibitem[{Edge et~al.(2024)Edge, Trinh, Cheng, Bradley, Chao, Mody, Truitt, and Larson}]{graphrag}
Darren Edge, Ha~Trinh, Newman Cheng, Joshua Bradley, Alex Chao, Apurva Mody, Steven Truitt, and Jonathan Larson. 2024.
\newblock \href {https://api.semanticscholar.org/CorpusID:269363075} {From local to global: A graph rag approach to query-focused summarization}.
\newblock \emph{ArXiv}, abs/2404.16130.

\bibitem[{Feng et~al.(2022)Feng, Bu, Zhang, and Li}]{feng2022beyond}
Weixin Feng, Xingyuan Bu, Chenchen Zhang, and Xubin Li. 2022.
\newblock Beyond bounding box: Multimodal knowledge learning for object detection.
\newblock \emph{arXiv preprint arXiv:2205.04072}.

\bibitem[{F{\'e}vry et~al.(2020)F{\'e}vry, Soares, FitzGerald, Choi, and Kwiatkowski}]{fevry2020entities}
Thibault F{\'e}vry, Livio~Baldini Soares, Nicholas FitzGerald, Eunsol Choi, and Tom Kwiatkowski. 2020.
\newblock Entities as experts: Sparse memory access with entity supervision.
\newblock \emph{arXiv preprint arXiv:2004.07202}.

\bibitem[{Fu et~al.(2024)Fu, Panda, Niu, Yue, Hajishirzi, Kim, and Peng}]{fu2024data}
Yao Fu, Rameswar Panda, Xinyao Niu, Xiang Yue, Hannaneh Hajishirzi, Yoon Kim, and Hao Peng. 2024.
\newblock Data engineering for scaling language models to 128k context.
\newblock \emph{arXiv preprint arXiv:2402.10171}.

\bibitem[{Gu and Dao(2023)}]{gu2023mamba}
Albert Gu and Tri Dao. 2023.
\newblock Mamba: Linear-time sequence modeling with selective state spaces.
\newblock \emph{arXiv preprint arXiv:2312.00752}.

\bibitem[{Ho et~al.(2020)Ho, Nguyen, Sugawara, and Aizawa}]{ho-2020-coling-2wikimultihopQA}
Xanh Ho, Anh{-}Khoa~Duong Nguyen, Saku Sugawara, and Akiko Aizawa. 2020.
\newblock Constructing {A} multi-hop {QA} dataset for comprehensive evaluation of reasoning steps.
\newblock In \emph{{COLING}}, pages 6609--6625. International Committee on Computational Linguistics.

\bibitem[{Jiang et~al.(2024)Jiang, Ma, and Chen}]{Jiang2024LongRAGER}
Ziyan Jiang, Xueguang Ma, and Wenhu Chen. 2024.
\newblock \href {https://api.semanticscholar.org/CorpusID:270688725} {Longrag: Enhancing retrieval-augmented generation with long-context llms}.
\newblock \emph{ArXiv}, abs/2406.15319.

\bibitem[{Khandelwal et~al.(2019)Khandelwal, Levy, Jurafsky, Zettlemoyer, and Lewis}]{khandelwal2019generalization}
Urvashi Khandelwal, Omer Levy, Dan Jurafsky, Luke Zettlemoyer, and Mike Lewis. 2019.
\newblock Generalization through memorization: Nearest neighbor language models.
\newblock \emph{arXiv preprint arXiv:1911.00172}.

\bibitem[{Khattab and Zaharia(2020)}]{khattab2020colbert}
Omar Khattab and Matei Zaharia. 2020.
\newblock Colbert: Efficient and effective passage search via contextualized late interaction over bert.
\newblock In \emph{Proceedings of the 43rd International ACM SIGIR conference on research and development in Information Retrieval}, pages 39--48.

\bibitem[{Kocisk{\'{y}} et~al.(2018)Kocisk{\'{y}}, Schwarz, Blunsom, Dyer, Hermann, Melis, and Grefenstette}]{Kocisky-2018-acltrans-narrativeqa}
Tom{\'{a}}s Kocisk{\'{y}}, Jonathan Schwarz, Phil Blunsom, Chris Dyer, Karl~Moritz Hermann, G{\'{a}}bor Melis, and Edward Grefenstette. 2018.
\newblock The narrativeqa reading comprehension challenge.
\newblock \emph{Trans. Assoc. Comput. Linguistics}, 6:317--328.

\bibitem[{Kwiatkowski et~al.(2019)Kwiatkowski, Palomaki, Redfield, Collins, Parikh, Alberti, Epstein, Polosukhin, Devlin, Lee, Toutanova, Jones, Kelcey, Chang, Dai, Uszkoreit, Le, and Petrov}]{kwiatkowski-etal-2019-natural}
Tom Kwiatkowski, Jennimaria Palomaki, Olivia Redfield, Michael Collins, Ankur Parikh, Chris Alberti, Danielle Epstein, Illia Polosukhin, Jacob Devlin, Kenton Lee, Kristina Toutanova, Llion Jones, Matthew Kelcey, Ming-Wei Chang, Andrew~M. Dai, Jakob Uszkoreit, Quoc Le, and Slav Petrov. 2019.
\newblock \href {https://doi.org/10.1162/tacl_a_00276} {Natural questions: A benchmark for question answering research}.
\newblock \emph{Transactions of the Association for Computational Linguistics}, 7:452--466.

\bibitem[{LangChain-team(2024)}]{LangChain}
LangChain-team. 2024.
\newblock \href {https://github.com/langchain-ai/langchain} {{LangChain}}.

\bibitem[{Lee et~al.(2024)Lee, Chen, Furuta, Canny, and Fischer}]{lee2024human}
Kuang-Huei Lee, Xinyun Chen, Hiroki Furuta, John Canny, and Ian Fischer. 2024.
\newblock A human-inspired reading agent with gist memory of very long contexts.
\newblock \emph{arXiv preprint arXiv:2402.09727}.

\bibitem[{Li et~al.(2023)Li, Zhao, Chia, Ding, Joty, Poria, and Bing}]{li2023chain}
Xingxuan Li, Ruochen Zhao, Yew~Ken Chia, Bosheng Ding, Shafiq Joty, Soujanya Poria, and Lidong Bing. 2023.
\newblock Chain-of-knowledge: Grounding large language models via dynamic knowledge adapting over heterogeneous sources.
\newblock In \emph{The Twelfth International Conference on Learning Representations}.

\bibitem[{Liu(2024)}]{LlamaIndex}
Jerry Liu. 2024.
\newblock \href {https://doi.org/10.5281/zenodo.1234} {{LlamaIndex}}.

\bibitem[{Liu et~al.(2024{\natexlab{a}})Liu, Zhou, Liu, Bu, Yang, Han-Sen, and Ouyang}]{liu2024iterative}
Jie Liu, Zhanhui Zhou, Jiaheng Liu, Xingyuan Bu, Chao Yang, Zhong Han-Sen, and Wanli Ouyang. 2024{\natexlab{a}}.
\newblock Iterative length-regularized direct preference optimization: A case study on improving 7b language models to gpt-4 level.
\newblock \emph{arXiv preprint arXiv:2406.11817}.

\bibitem[{Liu et~al.(2024{\natexlab{b}})Liu, Lin, Hewitt, Paranjape, Bevilacqua, Petroni, and Liang}]{liu2024lost}
Nelson~F Liu, Kevin Lin, John Hewitt, Ashwin Paranjape, Michele Bevilacqua, Fabio Petroni, and Percy Liang. 2024{\natexlab{b}}.
\newblock Lost in the middle: How language models use long contexts.
\newblock \emph{Transactions of the Association for Computational Linguistics}, 12:157--173.

\bibitem[{Lu et~al.(2023)Lu, Yuan, Yuan, Lin, Lin, Tan, Zhou, and Zhou}]{lu2023instag}
Keming Lu, Hongyi Yuan, Zheng Yuan, Runji Lin, Junyang Lin, Chuanqi Tan, Chang Zhou, and Jingren Zhou. 2023.
\newblock \# instag: Instruction tagging for analyzing supervised fine-tuning of large language models.
\newblock In \emph{The Twelfth International Conference on Learning Representations}.

\bibitem[{Luo et~al.(2023)Luo, Li, Haffari, and Pan}]{luo2023reasoning}
Linhao Luo, Yuan-Fang Li, Gholamreza Haffari, and Shirui Pan. 2023.
\newblock Reasoning on graphs: Faithful and interpretable large language model reasoning.
\newblock \emph{arXiv preprint arXiv:2310.01061}.

\bibitem[{Munkhdalai et~al.(2024)Munkhdalai, Faruqui, and Gopal}]{munkhdalai2024leave}
Tsendsuren Munkhdalai, Manaal Faruqui, and Siddharth Gopal. 2024.
\newblock Leave no context behind: Efficient infinite context transformers with infini-attention.
\newblock \emph{arXiv preprint arXiv:2404.07143}.

\bibitem[{Nakano et~al.(2021)Nakano, Hilton, Balaji, Wu, Ouyang, Kim, Hesse, Jain, Kosaraju, Saunders et~al.}]{nakano2021webgpt}
Reiichiro Nakano, Jacob Hilton, Suchir Balaji, Jeff Wu, Long Ouyang, Christina Kim, Christopher Hesse, Shantanu Jain, Vineet Kosaraju, William Saunders, et~al. 2021.
\newblock Webgpt: Browser-assisted question-answering with human feedback.
\newblock \emph{arXiv preprint arXiv:2112.09332}.

\bibitem[{Pang et~al.(2022)Pang, Parrish, Joshi, Nangia, Phang, Chen, Padmakumar, Ma, Thompson, He, and Bowman}]{pang-etal-2022-quality}
Richard~Yuanzhe Pang, Alicia Parrish, Nitish Joshi, Nikita Nangia, Jason Phang, Angelica Chen, Vishakh Padmakumar, Johnny Ma, Jana Thompson, He~He, and Samuel Bowman. 2022.
\newblock \href {https://doi.org/10.18653/v1/2022.naacl-main.391} {{Q}u{ALITY}: Question answering with long input texts, yes!}
\newblock In \emph{Proceedings of the 2022 Conference of the North American Chapter of the Association for Computational Linguistics: Human Language Technologies}, pages 5336--5358, Seattle, United States. Association for Computational Linguistics.

\bibitem[{Park et~al.(2023)Park, O'Brien, Cai, Morris, Liang, and Bernstein}]{park2023generative}
Joon~Sung Park, Joseph O'Brien, Carrie~Jun Cai, Meredith~Ringel Morris, Percy Liang, and Michael~S Bernstein. 2023.
\newblock Generative agents: Interactive simulacra of human behavior.
\newblock In \emph{Proceedings of the 36th Annual ACM Symposium on User Interface Software and Technology}, pages 1--22.

\bibitem[{Peng et~al.(2023{\natexlab{a}})Peng, Quesnelle, Fan, and Shippole}]{peng2023yarn}
Bowen Peng, Jeffrey Quesnelle, Honglu Fan, and Enrico Shippole. 2023{\natexlab{a}}.
\newblock Yarn: Efficient context window extension of large language models.
\newblock \emph{arXiv preprint arXiv:2309.00071}.

\bibitem[{Peng et~al.(2020)Peng, Bu, Sun, Zhang, Tan, and Yan}]{peng2020large}
Junran Peng, Xingyuan Bu, Ming Sun, Zhaoxiang Zhang, Tieniu Tan, and Junjie Yan. 2020.
\newblock Large-scale object detection in the wild from imbalanced multi-labels.
\newblock In \emph{Proceedings of the IEEE/CVF conference on computer vision and pattern recognition}, pages 9709--9718.

\bibitem[{Peng et~al.(2023{\natexlab{b}})Peng, Chang, Yin, Bu, Sun, Xie, Zhang, Tian, and Zhang}]{peng2023gaia}
Junran Peng, Qing Chang, Haoran Yin, Xingyuan Bu, Jiajun Sun, Lingxi Xie, Xiaopeng Zhang, Qi~Tian, and Zhaoxiang Zhang. 2023{\natexlab{b}}.
\newblock Gaia-universe: Everything is super-netify.
\newblock \emph{IEEE Transactions on Pattern Analysis and Machine Intelligence}, 45(10):11856--11868.

\bibitem[{Rasooli and Tetreault(2015)}]{rasooli-tetrault-2015}
Mohammad~Sadegh Rasooli and Joel~R. Tetreault. 2015.
\newblock \href {http://arxiv.org/abs/1503.06733} {Yara parser: {A} fast and accurate dependency parser}.
\newblock \emph{Computing Research Repository}, arXiv:1503.06733.
\newblock Version 2.

\bibitem[{Robertson and Zaragoza(2009)}]{Robertson2009ThePR}
Stephen~E. Robertson and Hugo Zaragoza. 2009.
\newblock \href {https://api.semanticscholar.org/CorpusID:207178704} {The probabilistic relevance framework: Bm25 and beyond}.
\newblock \emph{Found. Trends Inf. Retr.}, 3:333--389.

\bibitem[{Sachan et~al.(2023)Sachan, Lewis, Yogatama, Zettlemoyer, Pineau, and Zaheer}]{sachan2023questions}
Devendra~Singh Sachan, Mike Lewis, Dani Yogatama, Luke Zettlemoyer, Joelle Pineau, and Manzil Zaheer. 2023.
\newblock Questions are all you need to train a dense passage retriever.
\newblock \emph{Transactions of the Association for Computational Linguistics}, 11:600--616.

\bibitem[{Sarthi et~al.(2024)Sarthi, Abdullah, Tuli, Khanna, Goldie, and Manning}]{sarthi2024raptor}
Parth Sarthi, Salman Abdullah, Aditi Tuli, Shubh Khanna, Anna Goldie, and Christopher~D Manning. 2024.
\newblock Raptor: Recursive abstractive processing for tree-organized retrieval.
\newblock \emph{arXiv preprint arXiv:2401.18059}.

\bibitem[{Sun et~al.(2023)Sun, Xu, Tang, Wang, Lin, Gong, Shum, and Guo}]{sun2023think}
Jiashuo Sun, Chengjin Xu, Lumingyuan Tang, Saizhuo Wang, Chen Lin, Yeyun Gong, Heung-Yeung Shum, and Jian Guo. 2023.
\newblock Think-on-graph: Deep and responsible reasoning of large language model with knowledge graph.
\newblock \emph{arXiv preprint arXiv:2307.07697}.

\bibitem[{Sun et~al.(2021)Sun, Krishna, Mattarella-Micke, and Iyyer}]{sun2021long}
Simeng Sun, Kalpesh Krishna, Andrew Mattarella-Micke, and Mohit Iyyer. 2021.
\newblock Do long-range language models actually use long-range context?
\newblock \emph{arXiv preprint arXiv:2109.09115}.

\bibitem[{Sun et~al.(2024)Sun, Liu, Wang, Iter, Zhu, and Iyyer}]{sun-etal-2024-pearl}
Simeng Sun, Yang Liu, Shuohang Wang, Dan Iter, Chenguang Zhu, and Mohit Iyyer. 2024.
\newblock \href {https://aclanthology.org/2024.eacl-long.29} {{PEARL}: Prompting large language models to plan and execute actions over long documents}.
\newblock In \emph{Proceedings of the 18th Conference of the European Chapter of the Association for Computational Linguistics (Volume 1: Long Papers)}, pages 469--486, St. Julian{'}s, Malta. Association for Computational Linguistics.

\bibitem[{Trivedi et~al.(2022)Trivedi, Balasubramanian, Khot, and Sabharwal}]{Trivedi-2022-acltrans-musique}
Harsh Trivedi, Niranjan Balasubramanian, Tushar Khot, and Ashish Sabharwal. 2022.
\newblock Musique: Multihop questions via single-hop question composition.
\newblock \emph{Trans. Assoc. Comput. Linguistics}, 10:539--554.

\bibitem[{Wang et~al.(2024)Wang, Lipka, Rossi, Siu, Zhang, and Derr}]{wang2024knowledge}
Yu~Wang, Nedim Lipka, Ryan~A Rossi, Alexa Siu, Ruiyi Zhang, and Tyler Derr. 2024.
\newblock Knowledge graph prompting for multi-document question answering.
\newblock In \emph{Proceedings of the AAAI Conference on Artificial Intelligence}, volume~38, pages 19206--19214.

\bibitem[{Wei et~al.(2022)Wei, Wang, Schuurmans, Bosma, Xia, Chi, Le, Zhou et~al.}]{wei2022chain}
Jason Wei, Xuezhi Wang, Dale Schuurmans, Maarten Bosma, Fei Xia, Ed~Chi, Quoc~V Le, Denny Zhou, et~al. 2022.
\newblock Chain-of-thought prompting elicits reasoning in large language models.
\newblock \emph{Advances in neural information processing systems}, 35:24824--24837.

\bibitem[{Wu et~al.(2024{\natexlab{a}})Wu, Wang, Fu, Yue, Zhu, and Li}]{wu2024long}
Wenhao Wu, Yizhong Wang, Yao Fu, Xiang Yue, Dawei Zhu, and Sujian Li. 2024{\natexlab{a}}.
\newblock Long context alignment with short instructions and synthesized positions.
\newblock \emph{arXiv preprint arXiv:2405.03939}.

\bibitem[{Wu et~al.(2024{\natexlab{b}})Wu, Liu, Bu, Liu, Zhou, Zhang, Zhang, Bai, Chen, Ge et~al.}]{wu2024conceptmath}
Yanan Wu, Jie Liu, Xingyuan Bu, Jiaheng Liu, Zhanhui Zhou, Yuanxing Zhang, Chenchen Zhang, Zhiqi Bai, Haibin Chen, Tiezheng Ge, et~al. 2024{\natexlab{b}}.
\newblock Conceptmath: A bilingual concept-wise benchmark for measuring mathematical reasoning of large language models.
\newblock \emph{arXiv preprint arXiv:2402.14660}.

\bibitem[{Xv et~al.(2022)Xv, Chen, Lin, Guan, Bu, Li, Deng, Xu, and Zheng}]{xv2022visual}
Guipeng Xv, Si~Chen, Chen Lin, Wanxian Guan, Xingyuan Bu, Xubin Li, Hongbo Deng, Jian Xu, and Bo~Zheng. 2022.
\newblock Visual encoding and debiasing for ctr prediction.
\newblock In \emph{Proceedings of the 31st ACM International Conference on Information \& Knowledge Management}, pages 4615--4619.

\bibitem[{Yang et~al.(2018)Yang, Qi, Zhang, Bengio, Cohen, Salakhutdinov, and Manning}]{yang-2018-emnlp-hotpotqa}
Zhilin Yang, Peng Qi, Saizheng Zhang, Yoshua Bengio, William~W. Cohen, Ruslan Salakhutdinov, and Christopher~D. Manning. 2018.
\newblock Hotpotqa: {A} dataset for diverse, explainable multi-hop question answering.
\newblock In \emph{{EMNLP}}, pages 2369--2380. Association for Computational Linguistics.

\bibitem[{Yao et~al.(2022)Yao, Zhao, Yu, Du, Shafran, Narasimhan, and Cao}]{yao2022react}
Shunyu Yao, Jeffrey Zhao, Dian Yu, Nan Du, Izhak Shafran, Karthik Narasimhan, and Yuan Cao. 2022.
\newblock React: Synergizing reasoning and acting in language models.
\newblock \emph{arXiv preprint arXiv:2210.03629}.

\bibitem[{Yuan et~al.(2024)Yuan, Ning, Zhou, Yang, Li, Zhuang, Tan, Yao, Lin, Li, Dai, Yan, and Wang}]{Yuan2024LVEvalAB}
Tao Yuan, Xuefei Ning, Dong Zhou, Zhijie Yang, Shiyao Li, Minghui Zhuang, Zheyue Tan, Zhuyu Yao, Dahua Lin, Boxun Li, Guohao Dai, Shengen Yan, and Yu~Wang. 2024.
\newblock \href {https://api.semanticscholar.org/CorpusID:267547607} {Lv-eval: A balanced long-context benchmark with 5 length levels up to 256k}.
\newblock \emph{ArXiv}, abs/2402.05136.

\bibitem[{Zhao et~al.(2023)Zhao, Zhou, Li, Tang, Wang, Hou, Min, Zhang, Zhang, Dong, Du, Yang, Chen, Chen, Jiang, Ren, Li, Tang, Liu, Liu, Nie, and Wen}]{Zhao-2023-arxiv-survey}
Wayne~Xin Zhao, Kun Zhou, Junyi Li, Tianyi Tang, Xiaolei Wang, Yupeng Hou, Yingqian Min, Beichen Zhang, Junjie Zhang, Zican Dong, Yifan Du, Chen Yang, Yushuo Chen, Zhipeng Chen, Jinhao Jiang, Ruiyang Ren, Yifan Li, Xinyu Tang, Zikang Liu, Peiyu Liu, Jian{-}Yun Nie, and Ji{-}Rong Wen. 2023.
\newblock A survey of large language models.
\newblock abs/2303.18223.

\bibitem[{Zhu et~al.(2023)Zhu, Yang, Wang, Song, Wu, Wei, and Li}]{zhu2023pose}
Dawei Zhu, Nan Yang, Liang Wang, Yifan Song, Wenhao Wu, Furu Wei, and Sujian Li. 2023.
\newblock Pose: Efficient context window extension of llms via positional skip-wise training.
\newblock \emph{arXiv preprint arXiv:2309.10400}.

\end{thebibliography}
\clearpage
\appendix
\section{GraphReader Prompt}
\label{app: graphreader_prompt}
Figure~\ref{tab: extract_claims} illustrates the prompt used for Graph Construction.
Figures~\ref{tab: pre-planning} to~\ref{tab: read_neighbor} present the prompts employed for Graph Exploration.
Figure~\ref{tab: graph_summary} shows the prompt used for Answer Reasoning.




\begin{figure*}[t]
\begin{tcolorbox}[colback=yellow!6!white,colframe=blue!50!green]
You are now an intelligent assistant tasked with meticulously extracting both key elements and atomic facts from a long text.\\
1. Key Elements: The essential nouns (e.g., characters, times, events, places, numbers), verbs (e.g., actions), and adjectives (e.g., states, feelings) that are pivotal to the text's narrative.\\
2. Atomic Facts: The smallest, indivisible facts, presented as concise sentences. These include propositions, theories, existences, concepts, and implicit elements like logic, causality, event sequences, interpersonal relationships, timelines, etc.\\
\\
Requirements:\\
\#\#\#\#\#\\
1. Ensure that all identified key elements are reflected within the corresponding atomic facts.\\
2. You should extract key elements and atomic facts comprehensively, especially those that are important and potentially query-worthy and do not leave out details.\\
3. Whenever applicable, replace pronouns with their specific noun counterparts (e.g., change I, He, She to actual names).\\
4. Ensure that the key elements and atomic facts you extract are presented in the same language as the original text (e.g., English or Chinese).\\
5. You should output a total of key elements and atomic facts that do not exceed 1024 tokens.\\
6. Your answer format for each line should be: 
$[$Serial Number$]$, $[$Atomic Facts$]$, $[$List of Key Elements, separated with `|'$]$  \\
\#\#\#\#\#\\
\\
Example:\\
\#\#\#\#\#\\
User:\\
One day, a father and his little son ......\\
\\
Assistant:\\
1. One day, a father and his little son were going home. | father | little son | going home\\
2. ......\\
\#\#\#\#\#\\
\\
Please strictly follow the above format. Let's begin.
\end{tcolorbox}
\caption{The prompt for key elements and atomic facts extraction.}
\label{tab: extract_claims}
\end{figure*}
\begin{figure*}[t]
\begin{tcolorbox}[colback=yellow!6!white,colframe=blue!50!green]
As an intelligent assistant, your primary objective is to answer the question by gathering supporting facts from a given article. To facilitate this objective, the first step is to make a rational plan based on the question. This plan should outline the step-by-step process to resolve the question and specify the key information required to formulate a comprehensive answer.
\\
\\
Example:\\
\#\#\#\#\#\\
User: Who had a longer tennis career, Danny or Alice?\\
\\
Assistant: In order to answer this question, we first need to find the length of Danny's and Alice's tennis careers, such as the start and retirement of their careers, and then compare the two.\\
\#\#\#\#\#\\
\\
Please strictly follow the above format. Let's begin.
\end{tcolorbox}
\caption{The prompt for rational plan.}
\label{tab: pre-planning}
\end{figure*}
\begin{figure*}[t]
\begin{tcolorbox}[colback=yellow!6!white,colframe=blue!50!green]
As an intelligent assistant, your primary objective is to answer questions based on information contained within a text. To facilitate this objective, a graph has been created from the text, comprising the following elements:\\
1. Text Chunks: Chunks of the original text.\\
2. Atomic Facts: Smallest, indivisible truths extracted from text chunks.\\
3. Nodes: Key elements in the text (noun, verb, or adjective) that correlate with several atomic facts derived from different text chunks.\\
\\
Your current task is to check a list of nodes, with the objective of selecting the most relevant initial nodes from the graph to efficiently answer the question. You are given the question, the rational plan, and a list of node key elements. These initial nodes are crucial because they are the starting point for searching for relevant information.\\
\\
Requirements:\\
\#\#\#\#\#\\
1. Once you have selected a starting node, assess its relevance to the potential answer by assigning a score between 0 and 100. A score of 100 implies a high likelihood of relevance to the answer, whereas a score of 0 suggests minimal relevance.\\
2. Present each chosen starting node in a separate line, accompanied by its relevance score. Format each line as follows: Node: [Key Element of Node], Score: [Relevance Score].\\
3. Please select at least 10 starting nodes, ensuring they are non-repetitive and diverse.\\
4. In the user's input, each line constitutes a node. When selecting the starting node, please make your choice from those provided, and refrain from fabricating your own. The nodes you output must correspond exactly to the nodes given by the user, with identical wording.\\
\#\#\#\#\#\\
\\
Example: \\
\#\#\#\#\#\\
User:\\
Question: \{QUESTION\} \\
Plan: \{RATIONAL PLAN\}\\
Nodes: \{LIST OF KEY ELEMENTS\} \\

Assistant:\{LIST OF SELECTED NODES\} \\
\#\#\#\#\#\\
\\
Finally, I emphasize again that you need to select the starting node from the given Nodes, and it must be consistent with the words of the node you selected.
Please strictly follow the above format. Let's begin.
\end{tcolorbox}
\caption{The prompt for initial node selection.}
\label{tab: search_beginning_nodes}
\end{figure*}
\begin{figure*}[t]
\begin{tcolorbox}[colback=yellow!6!white,colframe=blue!50!green]
As an intelligent assistant, your primary objective is to answer questions based on information contained within a text. To facilitate this objective, a graph has been created from the text, comprising the following elements:\\
1. Text Chunks: Chunks of the original text.\\
2. Atomic Facts: Smallest, indivisible truths extracted from text chunks.\\
3. Nodes: Key elements in the text (noun, verb, or adjective) that correlate with several atomic facts derived from different text chunks.\\
\\
Your current task is to check a node and its associated atomic facts, with the objective of determining whether to proceed with reviewing the text chunk corresponding to these atomic facts. Given the question, the rational plan, previous actions, notebook content, and the current node's atomic facts and their corresponding chunk IDs, you have the following Action Options:\\
\#\#\#\#\#\\
1. read\_chunk(List[ID]): Choose this action if you believe that a text chunk linked to an atomic fact may hold the necessary information to answer the question. This will allow you to access more complete and detailed information.\\
2. stop\_and\_read\_neighbor(): Choose this action if you ascertain that all text chunks lack valuable information.\\
\#\#\#\#\#\\
\\
Strategy:\\
\#\#\#\#\#\\
1. Reflect on previous actions and prevent redundant revisiting nodes or chunks.\\
2. You can choose to read multiple text chunks at the same time.\\
3. Atomic facts only cover part of the information in the text chunk, so even if you feel that the atomic facts are slightly relevant to the question, please try to read the text chunk to get more complete information.\\
\#\#\#\#\#\\
\\
Response format:\\
\#\#\#\#\#\\
*Updated Notebook*: First, combine your current notebook with new insights and findings about the question from current atomic facts, creating a more complete version of the notebook that contains more valid information.\\
*Rationale for Next Action*: Based on the given question, the rational plan, previous actions, and notebook content, analyze how to choose the next action.\\
*Chosen Action*: read\_chunk(List[ID]) or stop\_and\_read\_neighbor(). (Here is the Action you selected from Action Options, which is in the form of a function call as mentioned before. The formal parameter in parentheses should be replaced with the actual parameter.)\\
\#\#\#\#\#\\
\\
Finally, it is emphasized again that even if the atomic fact is only slightly relevant to the question, you should still look at the text chunk to avoid missing information. You should only choose stop\_and\_read\_neighbor() when you are very sure that the given text chunk is irrelevant to the question.
Please strictly follow the above format. Let's begin.
\end{tcolorbox}
\caption{The prompt for exploring atomic facts.}
\label{tab: read_atomic_facts}
\end{figure*}
\begin{figure*}[t]
\begin{tcolorbox}[colback=yellow!6!white,colframe=blue!50!green]
As an intelligent assistant, your primary objective is to answer questions based on information within a text. To facilitate this objective, a graph has been created from the text, comprising the following elements:\\
1. Text Chunks: Segments of the original text.\\
2. Atomic Facts: Smallest, indivisible truths extracted from text chunks.\\
3. Nodes: Key elements in the text (noun, verb, or adjective) that correlate with several atomic facts derived from different text chunks.\\
\\

Your current task is to assess a specific text chunk and determine whether the available information suffices to answer the question. Given the question, rational plan, previous actions, notebook content, and the current text chunk, you have the following Action Options:\\
\#\#\#\#\#\\
1. search\_more(): 
Choose this action if you think that the essential information necessary to answer the question is still lacking. 
\\
2. read\_previous\_chunk(): Choose this action if you feel that the previous text chunk contains valuable information for answering the question.
\\
3. read\_subsequent\_chunk(): Choose this action if you feel that the subsequent text chunk contains valuable information for answering the question.
\\
4. termination(): Choose this action if you believe that the information you have currently obtained is enough to answer the question. This will allow you to summarize the gathered information and provide a final answer.
\\
\#\#\#\#\#\\

Strategy:\\
\#\#\#\#\#\\
1. Reflect on previous actions and prevent redundant revisiting of nodes or chunks.\\
2. You can only choose one action.\\
\#\#\#\#\#\\

Response format:\\
\#\#\#\#\#\\
*Updated Notebook*: First, combine your previous notes with new insights and findings about the question from current text chunks, creating a more complete version of the notebook that contains more valid information.\\
*Rationale for Next Action*: Based on the given question, rational plan, previous actions, and notebook content, analyze how to choose the next action.\\
*Chosen Action*: search\_more() or read\_previous\_chunk() or read\_subsequent\_chunk() or termination(). (Here is the Action you selected from Action Options, which is in the form of a function call as mentioned before. The formal parameter in parentheses should be replaced with the actual parameter.)\\
\#\#\#\#\#\\
\\
Please strictly follow the above format. Let's begin.
\end{tcolorbox}
\caption{The prompt for exploring chunks.}
\label{tab: read_text_chunk}
\end{figure*}
\begin{figure*}[t]
\begin{tcolorbox}[colback=yellow!6!white,colframe=blue!50!green]
As an intelligent assistant, your primary objective is to answer questions based on information within a text. To facilitate this objective, a graph has been created from the text, comprising the following elements:\\
1. Text Chunks: Segments of the original text.\\
2. Atomic Facts: Smallest, indivisible truths extracted from text chunks.\\
3. Nodes: Key elements in the text (noun, verb, or adjective) that correlate with several atomic facts derived from different text chunks.\\
\\
Your current task is to assess all neighboring nodes of the current node, with the objective of determining whether to proceed to the next neighboring node.
Given the question, rational plan, previous actions, notebook content, and the neighbors of the current node, you have the following Action Options:\\
\#\#\#\#\#\\
1. read\_neighbor\_node(key element of node): Choose this action if you believe that any of the neighboring nodes may contain information relevant to the question. Note that you should focus on one neighbor node at a time.
\\
2. termination(): Choose this action if you believe that none of the neighboring nodes possess information that could answer the question.\\
\#\#\#\#\#\\

Strategy:\\
\#\#\#\#\#\\
1. Reflect on previous actions and prevent redundant revisiting of nodes or chunks.\\
2. You can only choose one action. This means that you can choose to read only one neighbor node or choose to terminate.\\
\#\#\#\#\#\\

Response format:\\
\#\#\#\#\#\\
*Rationale for Next Action*: Based on the given question, rational plan, previous actions, and notebook content, analyze how to choose the next action.\\
*Chosen Action*: read\_neighbor\_node(neighbor\_node) or termination(). (Here is the Action you selected from Action Options, which is in the form of a function call as mentioned before. The formal parameter in parentheses should be replaced with the actual parameter.)\\
\#\#\#\#\#\\
\\
Please strictly follow the above format. Let's begin.\\
\end{tcolorbox}
\caption{The prompt for exploring neighbors.}
\label{tab: read_neighbor}
\end{figure*}
\begin{figure*}[t]
\begin{tcolorbox}[colback=yellow!6!white,colframe=blue!50!green]
As an intelligent assistant, your primary objective is to answer questions based on information within a text. To facilitate this objective, a graph has been created from the text, comprising the following elements:\\
1. Text Chunks: Segments of the original text.\\
2. Atomic Facts: Smallest, indivisible truths extracted from text chunks.\\
3. Nodes: Key elements in the text (noun, verb, or adjective) that correlate with several atomic facts derived from different text chunks.\\
\\
You have now explored multiple paths from various starting nodes on this graph, recording key information for each path in a notebook. \\
Your task now is to analyze these memories and reason to answer the question.\\

Strategy:\\
\#\#\#\#\#\\
1. You should first analyze each notebook content before providing a final answer.\\
2. During the analysis, consider complementary information from other notes and employ a majority voting strategy to resolve any inconsistencies.\\
3. When generating the final answer, ensure that you take into account all available information.\\
\#\#\#\#\#\\

Example:\\
\#\#\#\#\#\\
User:\\
Question: Who had a longer tennis career, Danny or Alice?
\\
Notebook of different exploration paths:\\
1. We only know that Danny's tennis career started in 1972 and ended in 1990, but we don't know the length of Alice's career.\\
2. ......\\

Assistant:\\
Analyze:\\
The summary of search path 1 points out that Danny's tennis career is 1990-1972=18 years. 
Although it does not indicate the length of Alice's career, the summary of search path 2 finds this information, that is, the length of Alice's tennis career is 15 years.
Then we can get the final answer, that is, Danny's tennis career is longer than Alice's.\\
Final answer:\\
Danny's tennis career is longer than Alice's.\\
\#\#\#\#\#\\
\\
Please strictly follow the above format. Let's begin.\\
\end{tcolorbox}
\caption{The prompt for answer reasoning.}
\label{tab: graph_summary}
\end{figure*}

\section{LLM Rater Evaluation Details}
\label{app: eval_details}

Given a question, a golden answer, and an answer to be evaluated, we utilize an LLM to assess the accuracy of the latter based on the question and correct answer. This involves two scores: LLM-Rating-1 (LR-1) and LLM-Rating-2 (LR-2), where LR-1 represents a strict scoring criterion, and LR-2 is a more lenient one. Following the approach of ReadAgent, if either LLM Rater deems an answer correct, it is considered as such. If the strict scorer finds an answer incorrect while the lenient scorer deems it partially correct, we classify the answer as partially correct; otherwise, it is adjudged incorrect. The prompts used for evaluation are presented in Figure \ref{fig: lr1_prompt} and Figure \ref{fig: lr2_prompt} respectively.

For the evaluation, we utilize GPT-4-128k as the LLM Rater, with the temperature set to 0.1.  
\begin{figure*}[htbp]
\begin{tcolorbox}[colback=yellow!6!white,colframe=blue!50!green]
After reading some text, John was given the following question about the text: \\
\{QUESTION TEXT\}\\
John’s answer to the question was:\\
\{MODEL RESPONSE TEXT\}\\
The ground truth answer was: \\
\{REFERENCE RESPONSE TEXT\}\\
\\
Does John’s answer agree with the ground truth answer? \\
Please answer "Yes" or "No".
\end{tcolorbox}
\caption{The prompt for LLM-Rating-1.}
\label{fig: lr1_prompt}
\end{figure*}
\begin{figure*}[htbp]
\begin{tcolorbox}[colback=yellow!6!white,colframe=blue!50!green]
After reading some text, John was given the following question about the text:      \\
\{QUESTION TEXT\}\\
John’s answer to the question was: \\
\{MODEL RESPONSE TEXT\}\\
The ground truth answer was: \\
\{REFERENCE RESPONSE TEXT\}\\
\\
Does John’s answer agree with the ground truth answer?\\
Please answer “Yes”, “Yes, partially”, or “No”. If John’s response has any overlap with the ground truth answer, answer “Yes, partially”. If John’s response contains the ground truth answer, answer “Yes”. If John’s response is more specific than the ground truth answer, answer “Yes”.
\end{tcolorbox}
\caption{The prompt for LLM-Rating-2.}
\label{fig: lr2_prompt}
\end{figure*}
\section{Dataset}
\label{app: datasets}

\paragraph{Multi-hop QA Datasets} 
\textbf{HotpotQA} features a collection of 2-hop questions directly authored by native speakers, based on two interconnected paragraphs. \textbf{2WikiMultihopQA} is comprised of complex questions up to 5-hops in length, constructed through carefully designed templates to prevent the possibility of shortcut solutions.

In the \textbf{MuSiQue} dataset, questions are intricately crafted starting from straightforward scenarios that require up to 4-hops reasoning. 
Annotators subsequently rephrase these with a dual purpose: to avoid shortcut answers and to maintain a natural linguistic quality. 
Each question within the original datasets is complemented by 2-4 supporting paragraphs, delivering evidence for simple one-step reasoning, alongside multiple paragraphs designed to serve as decoys.

\textbf{HotpotWikiQA-mixup} originates from LV-Eval and employs a construction method known as a mixup. This method randomly blends support documents with various distracting documents to generate five different context lengths for a given QA pair, including 16k, 32k, 64k, 128k, and 256k. Due to the excessive length of this dataset, we select the first 50 data entries from each different context length for experimentation to control costs.

\paragraph{Single-hop QA Datasets}  
\textbf{NarrativeQA} is a dataset designed to test comprehension abilities for long documents, primarily sourced from movie scripts. As a single-hop QA dataset, the information required to answer its questions appears at a single location within the text.

\paragraph{Real-World Datasets}

\textbf{QuALITY}~\citep{pang-etal-2022-quality} is a long-text multiple-choice question-answering dataset, with questions crafted by contributors who are familiar with the complete passages, making it more representative of real-world QA scenarios. We handle it as straightforward QA problems. \textbf{Natural Questions}~\citep{kwiatkowski-etal-2019-natural} includes real anonymous aggregated queries from Google along with corresponding Wikipedia pages, providing another excellent resource for authentic long-text QA situations.
\vspace{-0.1cm}
\section{Baseline Methods}
\label{app: baseline}
\vspace{-0.1cm}

\paragraph{Full or Chunked Text Content} For texts with fewer tokens than the LLM's input window, we can input the text directly into the LLM to obtain an answer. We refer to this method as \textbf{Full Text Read}, with the specific prompt provided in Figure \ref{fig: overall_read_prompts}. However, this approach is not applicable to texts exceeding the token limit of the LLM's input window. In such cases, \citeauthor{lee2024human} truncated the text to fit it into the LLM, but this method obviously results in information loss. We propose a method that does not lose information, offering a better comparison. This method involves dividing the entire text into chunks (using the same chunking method as \ourmethod) and then having the LLM read these chunks sequentially according to the text order, thus enabling the handling of overly long texts with a limited input window. During the reading process, there are two main strategies: \textbf{Chunk Read} and \textbf{Chunk Read with Notes}. In the Chunk Read approach, the LLM only sees the current chunk during each reading, which is suitable for single-hop QA tasks. In the Chunk Read with Notes approach, the LLM can summarize useful information from the current chunk and provide it to the subsequent reading process, which is suitable for multi-hop QA tasks. 

In the experiment, we divide the chunks in the same way as \ourmethod, and the maximum length of the chunk is set to 2k. The specific prompts are in Figure \ref{fig: chunk_read_prompts} and \ref{fig: chunk_read_with_cache_prompts} respectively.

\begin{figure*}[!h]
\begin{tcolorbox}[colback=yellow!6!white,colframe=blue!50!green]
Please read the passage below and answer the question based on the passage.\\
Passage: \\
\{PASSAGE TEXT\}\\
Question: \\
\{QUESTION TEXT\}\\
\\
Now please answer this question based on the passage content.
\end{tcolorbox}
\caption{The prompt for Full Text Read.}
\label{fig: overall_read_prompts}
\end{figure*}
\begin{figure*}[htbp]
\begin{tcolorbox}[colback=yellow!6!white,colframe=blue!50!green]
Please read the text chunks below and answer the question.\\
Text chunks: \\
\{CHUNKED PASSAGE TEXT\}\\
Question: \\
\{QUESTION TEXT\}\\
\\
If you think you can answer the question based on the above text chunks please output [answerable] and then output your answer.\\
Otherwise, if there is not enough information to answer the question, please output:\\
{[}unanswerable]
\end{tcolorbox}
\caption{The prompt for Chunk Read.}
\label{fig: chunk_read_prompts}
\end{figure*}
\begin{figure*}[htbp]
\begin{tcolorbox}[colback=yellow!6!white,colframe=blue!50!green]
Please read the text chunk below and answer the questions based on your previous summary.\\
Text chunk:\\
\{CHUNKED PASSAGE TEXT\}\\
Your previous summary: \\
\{SUMMARY TEXT\}\\
Question: \\
\{QUESTION TEXT\}\\
\\
If the above text chunk has information that can help answer the question, please extract the effective information, output [summary], and then output the refined information. Please note that it must be brief.\\
If you can answer the question based on the above information, please output [answerable] and then output your answer.\\
Otherwise, if there is insufficient information to answer the question, please output {[}unanswerable].\\
\end{tcolorbox}
\caption{The prompt for Chunk Read with Note.}
\label{fig: chunk_read_with_cache_prompts}
\end{figure*}


\paragraph{Retrieval-Augmented Generation (RAG)} RAG is a commonly used approach for addressing long-text problems. In this work, we compare the traditional RAG method, including retrieval methods based on Okapi BM25~\citep{Robertson2009ThePR} and the OpenAI API embedding model (text-embedding-ada-002). Specifically, we first split the text into chunks in the same method as \ourmethod, then use the aforementioned methods to calculate the relevance scores between the question and these chunks, and finally input the \textit{top-n} chunks with the highest relevance scores together with the question for the LLM to answer. To ensure a fair comparison, we control the input window to 4k in the experiments. Specifically, in order to fill the input window as much as possible, we set the maximum length of the chunk to 38k when selecting the \textit{top-1} chunk for answering; when opting for the \textit{top-3} chunks, we set the maximum length of each chunk to 1k. The specific prompt can be found in  Figure \ref{fig: rag_prompts}.

In addition to traditional RAG methods, we also compared GraphRAG~\citep{graphrag} and LongRAG~\citep{Jiang2024LongRAGER}. GraphRAG utilizes LLMs to construct a graph-based text index in two distinct stages. The first stage involves extracting an entity knowledge graph from the source documents, while the second stage focuses on generating summaries for groups of entities. When a question is posed, each summary provides partial information, which is then combined into the final answer for the user. LongRAG introduces a ``long retriever'' and a ``long reader'', allowing the entire corpus to be processed into larger-sized units, which reduces the number of units needed during retrieval and alleviates the burden on the retriever.
\begin{figure*}[!htb]
\begin{tcolorbox}[colback=yellow!6!white,colframe=blue!50!green]
Please read the text chunk below and answer the question.\\
Text chunks: \\
\{RETRIEVED PASSAGE TEXT\}\\
Question: \\
\{QUESTION TEXT\}\\
\\
Now please answer this question based on the text chunks.
\end{tcolorbox}
\caption{The prompt for RAG.}
\label{fig: rag_prompts}
\end{figure*}


\paragraph{Agent Style Methods} We also compared our method with similar approaches for handling long texts with small input windows, such as ReadAgent~\citep{lee2024human}. ReadAgent is a method that segments long texts and generates gist memories, which are then looked up to search for information in order to answer questions. 
In the experiments, for datasets from LongBench, we adopted the default hyperparameters declared in the ReadAgent paper, specifically a max\_words of 600 and min\_words of 280 when splitting pages. For HotpotWikiQA-mixup from LV-Eval, we scaled these two hyperparameters using the same approach as in the ReadAgent paper. Specifically, for datasets with lengths of 256k and 128k, we used max\_words=10000 and min\_words=2000; for those with lengths of 64k, 32k and 16k, we used max\_words=5000 and min\_words=1000. At the same time, we employed the ReadAgent-S method, which ReadAgent claims to be the most effective, reading the pages in sequence. Additionally, we allowed reading up to 5 pages (Look up 1-5 pages).

We also compared PEARL~\citep{sun-etal-2024-pearl}, a prompting framework to enhance reasoning capabilities for long documents. PEARL is structured into three stages: action mining, plan formulation, and plan execution. It decomposes complex questions into actionable steps and utilizes LLMs for zero-shot or few-shot prompting execution.


\section{Additional Experimental Results}
\label{app: addtional_exp}

\vspace{-0.1cm}
\begin{table}[htbp]
\renewcommand{\arraystretch}{1.1}
\centering
\resizebox{\columnwidth}{!}{
    \begin{tabular}{lcrrrr|rrrr}
    \toprule
    \multirow{2}{*}{\textbf{Method}} & \textbf{Input}  & \multicolumn{4}{c}{\textbf{QuALITY}} & \multicolumn{4}{c}{\textbf{Natural Question}} \\ \cline{3-10}
     & \textbf{Window} & LR-1 & LR-2 & EM & $F_1$ & LR-1 & LR-2 & EM & $F_1$ \\ \midrule
    GPT-4-128k & $128k$ & 45.7 & 60.3 & 2.7 & 9.9 & 75.0 & 81.0 & 41.0 & 57.2 \\
    Pearl & $128k$ & \underline{52.3} & \underline{72.7} & 3.0 & 9.6 & 74.0 & 79.0 & 38.0 & 56.8 \\
    LongRAG & $128k$ & 52.3 & 67.0 & \underline{3.7} & \underline{11.6} & \underline{77.7} & \underline{83.0} & \underline{47.3} & \underline{60.0} \\
    GraphRAG & $128k$ & 46.0 & 68.7 & 2.0 & 6.1 & 67.0 & 77.0 & 47.0 & 53.2 \\
    \ourmethod & $4k$ & \textbf{57.3} & \textbf{82.3} & \textbf{4.3} & \textbf{14.3} & \textbf{79.0} & \textbf{84.7} & \textbf{48.3} & \textbf{62.1} \\
    \bottomrule
    \end{tabular}
}
\vspace{-0.1cm}
\caption{Performance~(\%) comparison of different baselines on two additional datasets. The best performance and the second-best performance are denoted in bold and underlined fonts, respectively.}
\label{tab: addtional_result}
\end{table}
\vspace{-0.1cm}

Table \ref{tab: addtional_result} presents additional experimental results for two datasets, QuALITY and Natural Questions, both of which are highly relevant to real-world question-answering scenarios. The results indicate that our method significantly outperforms other baseline models in real-world scenarios.

\section{Evaluation Recall for Supporting Facts}
\label{app: recall}
We evaluate the recall rate of supporting facts for different methods using GPT-4-128k, with the temperature set to $0.1$. Figure \ref{fig: recall_prompt} shows the specific evaluation prompt.

For \ourmethod, we evaluate the memory recorded in the final notebook. For ReadAgent, the evaluation focused on the final text segments reviewed. In the case of Chunk Read with Notes, we evaluate both the memory and the chunk read at the time of the final answer; for the RAG methods, we assess the retrieved chunks.
\begin{figure*}[htbp]
\begin{tcolorbox}[colback=yellow!6!white,colframe=blue!50!green]
Now you are an intelligent assistant. Given a text, a question, and $x$ supporting facts that can answer the question, please determine how many supporting facts the text covers.\\
\\
Requirements:\\
\#\#\#\#\#\\
1. It's possible that not all supporting facts are needed to answer the question, so you'll need to analyze the supporting facts to determine which supporting facts are actually needed, and then determine whether those supporting facts are covered. Supporting facts that are not really needed are discarded, and you do not need to judge whether they are covered. So the number of real supporting facts is t (0 < t <= x).\\
2. A supporting fact has some valid information that helps answer the question. When the text provides this part of the valid information, it is considered to have covered the supporting fact, even if the text does not provide all the information supporting the fact.\\
3. The number of covered items in your output should be between 0 and t (including 0 and t).\\
4. Please analyze and reason first, and then output the final result.\\
\#\#\#\#\#\\
\\
Example: \\
\#\#\#\#\#\\
\{EXAMPLE\}\\
\#\#\#\#\#\\
\\
Please note that you should follow: 0 <= Number of recalls <= True number of supporting facts <= Number of supporting facts.\\
Please output according to the example format. Now let's start.
\end{tcolorbox}
\caption{The prompt for evaluating recall.}
\label{fig: recall_prompt}
\end{figure*}

\section{The Analysis of Function Calls}
\label{app: func_call}

To verify the rationality and utility of agent actions under various circumstances of \ourmethod, we made statistics on its function calls at each stage across two datasets. 
From the statistical results in Table \ref{tab: action}, it can be observed that each piece of data will perform an average of 3 to 4 actions, corresponding to the average number of function calls in the table. This indicates the effectiveness of the graph we constructed, with \ourmethod being able to swiftly locate key information while minimizing resource usage. 
Furthermore, each action has a certain probability of being chosen, justifying the rationality of the action set. Among them, the most commonly used action on multi-hop QA tasks is to read neighbor nodes, and the most common action on single-hop QA tasks is to read chunks. This difference is caused by the fact that multi-hop questions need to gather information contained by multiple nodes to answer questions, while single-hop data sets often require only one atomic fact.

\begin{table*}[!t]
\centering
\resizebox{\textwidth}{!}{
    \begin{tabular}{cccccr}
    \toprule
    \textbf{Dataset} & \textbf{\#Avg. Function Call} & \textbf{Stage} & \textbf{Stage Ratio(\%)} & \textbf{Function} & \textbf{Call Ratio(\%)}\\ \hline

    \multirow{7}{*}{HotpotQA} & \multirow{7}{*}{3.0} & \multirow{2}{*}{Exploring Atomic Facts} & \multirow{2}{*}{42.0} & \textit{read\_chunk} & 46.5 \\ 
    &  &  &  & \textit{stop\_and\_read\_neighbor} & 53.5 \\ \cline{3-6} 
    & & \multirow{4}{*}{Exploring Chunks}     & \multirow{4}{*}{31.9} & \textit{search\_more} & 12.1  \\
    &  &  &  & \textit{read\_previous\_chunk} & 21.1 \\ 
    &  &  &  & \textit{read\_subsequent\_chunk} & 22.9 \\
    &  &  &  & \textit{termination} & 43.9 \\  \cline{3-6}
    &  & \multirow{2}{*}{Exploring Neighbors} & \multirow{2}{*}{26.1} & \textit{read\_neighbor\_node} & 35.5 \\
    &  &  &  & \textit{termination} & 65.5\\ \hline

    \multirow{7}{*}{2WikiMultihopQA} & \multirow{7}{*}{3.2} & \multirow{2}{*}{Exploring Atomic Facts} & \multirow{2}{*}{40.4} & \textit{read\_chunk} & 48.6 \\  
    &  &  &  & \textit{stop\_and\_read\_neighbor} & 51.4 \\ \cline{3-6} 
    & & \multirow{4}{*}{Exploring Chunks} & \multirow{4}{*}{34.5} & \textit{search\_more} & 14.5 \\
    &  &  &  & \textit{read\_previous\_chunk} & 25.1 \\
    &  &  &  & \textit{read\_subsequent\_chunk} & 23.3 \\
    &  &  &  & \textit{termination} & 37.1 \\ 
    \cline{3-6} 
    &  & \multirow{2}{*}{Exploring Neighbors} & \multirow{2}{*}{25.1} & \textit{read\_neighbor\_node} & 37.3  \\
    &  &  &  & \textit{termination} & 62.7  \\\hline
 
     \multirow{7}{*}{MuSiQue} & \multirow{7}{*}{3.5} & \multirow{2}{*}{Exploring Atomic Facts} & \multirow{2}{*}{40.0} & \textit{read\_chunk} & 41.3 \\ 
    &  &  &  & \textit{stop\_and\_read\_neighbor} & 58.7 \\ \cline{3-6} 
    & & \multirow{4}{*}{Exploring Chunks}     & \multirow{4}{*}{31.2} & \textit{search\_more} & 19.1  \\
    &  &  &  & \textit{read\_previous\_chunk} & 26.6 \\ 
    &  &  &  & \textit{read\_subsequent\_chunk} & 25.7 \\
    &  &  &  & \textit{termination} & 28.6 \\  \cline{3-6}
    &  & \multirow{2}{*}{Exploring Neighbors} & \multirow{2}{*}{28.8} & \textit{read\_neighbor\_node} & 40.1 \\
    &  &  &  & \textit{termination} & 59.9\\ \hline
\multirow{7}{*}{NarrativeQA} & \multirow{7}{*}{3.9} & \multirow{2}{*}{Exploring Atomic Facts} & \multirow{2}{*}{32.5} & \textit{read\_chunk} & 64.5 \\  
    &  &  &  & \textit{stop\_and\_read\_neighbor} & 35.5 \\ \cline{3-6} 
    & & \multirow{4}{*}{Exploring Chunks} & \multirow{4}{*}{54.3} & \textit{search\_more} & 4.1 \\
    &  &  &  & \textit{read\_previous\_chunk} & 35.3 \\
    &  &  &  & \textit{read\_subsequent\_chunk} & 32.6 \\
    &  &  &  & \textit{termination} & 28.0 \\ 
    \cline{3-6} 
    &  & \multirow{2}{*}{Exploring Neighbors} & \multirow{2}{*}{13.2} & \textit{read\_neighbor\_node} & 51.4  \\
    &  &  &  & \textit{termination} & 48.6  \\
    
    \bottomrule
    \end{tabular}
}
\caption{Statistics of function calls on MuSiQue and NarrativeQA.}
\label{tab: action}
\end{table*}

\vspace{-0.1cm}
\section{Statistics of Graph}
\label{app: statistical_data}
\vspace{-0.1cm}
The statistics of graphs from various datasets are presented in Table \ref{tab: graph}. For longer texts, there tends to be a higher average number of nodes and atomic facts. After normalization, each node has an average of about 10 neighbor nodes. This is because the number of key elements occurring simultaneously in each atomic fact is generally of this magnitude. Furthermore, the aggregation of similar nodes caused by normalization results in a slight increase in the number of neighboring nodes.

On average, each node is associated with about 2 atomic facts, and the average number of atomic facts in the node with the most atomic facts in each graph ranges from 15 to 50, indicating a relatively even distribution of atomic facts. The maximum average number of atomic facts is found in NarrativeQA, a possible explanation being that NarrativeQA is mainly derived from movie scripts, where characters, such as the protagonist, appear frequently throughout the text, thus including a larger number of atomic facts.

\begin{table*}[!t]
\centering
\resizebox{\textwidth}{!}{

\begin{tabular}{ccrrrr|rrrr}
\bottomrule
\multicolumn{2}{c}{\multirow{3}{*}{dataset}} & \multicolumn{4}{c}{Sample Dimension}                                & \multicolumn{4}{c}{Sample \& Node Dimension}                                 \\\cline{3-10}
\multicolumn{2}{c}{}                         & \multicolumn{2}{c}{node num} & \multicolumn{2}{c}{atomic facts num} & \multicolumn{2}{l}{neighbor node num} & \multicolumn{2}{l}{atomic facts num} \\\cline{3-10}
\multicolumn{2}{c}{}                         & avg.           & max         & avg.               & max             & avg. avg.             & avg. max               & avg. avg.            & avg. max               \\\hline
\multicolumn{2}{l}{HotpotQA}                 & 583.8          & 1945.0        & 244.0                & 645.0            & 10.1           & 263.1             & 2.1            & 17.8           \\
\multicolumn{2}{l}{2WikiMultihopQA}          & 515.8         & 1691.0        & 217.7             & 545.0           & 9.2             & 215.7            & 2.1             & 17.0                \\
\multicolumn{2}{l}{MusiQue}                  & 1029.4        & 2142.0        & 419.9            & 586.0             & 9.3              & 253.4             & 2.1              & 15.6              \\
\multicolumn{2}{l}{NarrativeQA}              & 966.0         & 3110.0       & 515.5             & 1296.0            & 10.3            & 652.6             & 2.3             & 50.0                \\\hline
\multirow{5}{*}{HotpotWikiQA-mixup}  & 16k   & 1741.6         & 3822.0        & 749.7             & 1043.0            & 9.4              & 231.0               & 2.2             & 17.1              \\
                                     & 32k   & 2827.3       & 5086.0        & 1257.4            & 1694.0            & 9.8              & 263.3             & 2.2             & 29.3              \\
                                     & 64k   & 5054.1        & 8918.0        & 2360.0               & 3015.0            & 10.4             & 227.2             & 2.3              & 17.1              \\
                                     & 128k  & 8828.5        & 14592.0       & 4437.9             & 5182.0            & 11.1             & 302.0               & 2.4             & 19.2              \\
                                     & 256k  & 14853.3        & 24981.0       & 8632.8             & 9478.0            & 12.2             & 427.6             & 2.5             & 27.8             \\
    \bottomrule
    \end{tabular}
}

\caption{Graph statistical data. Under the Sample dimension, \textbf{``avg.''} indicates the average number of nodes in each graph, and \textbf{``max''} refers to the largest node count across all graphs. The same logic applies to atomic facts num. In the Sample \& Node dimensions, \textbf{``avg. avg.''} denotes the average of the average neighbor node counts per graph, and \textbf{``avg. max''} means the average of the maximum neighbor node counts per graph. This approach is also used for counting atomic facts num.}
\label{tab: graph}
\end{table*}
\vspace{-0.1cm}
\section{\ourmethod Example}
\label{app: case}
\begin{figure*}[!htb]
\begin{tcolorbox}[colback=white,colframe=blue!50!green,title=\textbf{Graph Construction: Extract Atomic Facts And Key Elements}, center title]

\textbf{Atomic Facts}\\
1. "Never Too Loud" is the fourth studio album by Canadian hard rock band Danko Jones.\\
2. Danko Jones is a Canadian hard rock trio from Toronto.\\
3. Casa Loma is a Gothic Revival castle-style mansion and garden in midtown Toronto, Ontario, Canada.\\
......\\
\\
\textbf{Key Elements}\\
1. {[Never Too Loud, studio album, Canadian, hard rock band, Danko Jones]}\\
2. {[Danko Jones, Canadian, hard rock trio, Toronto]}\\
3. {[Casa Loma, Gothic Revival, castle-style mansion, Toronto, Canada]}\\
......
\end{tcolorbox}

\begin{tcolorbox}[colback=white,colframe=blue!50!green,title=\textbf{Graph Construction: Normalize And Link Nodes}, center title]
\includegraphics[width=\textwidth]{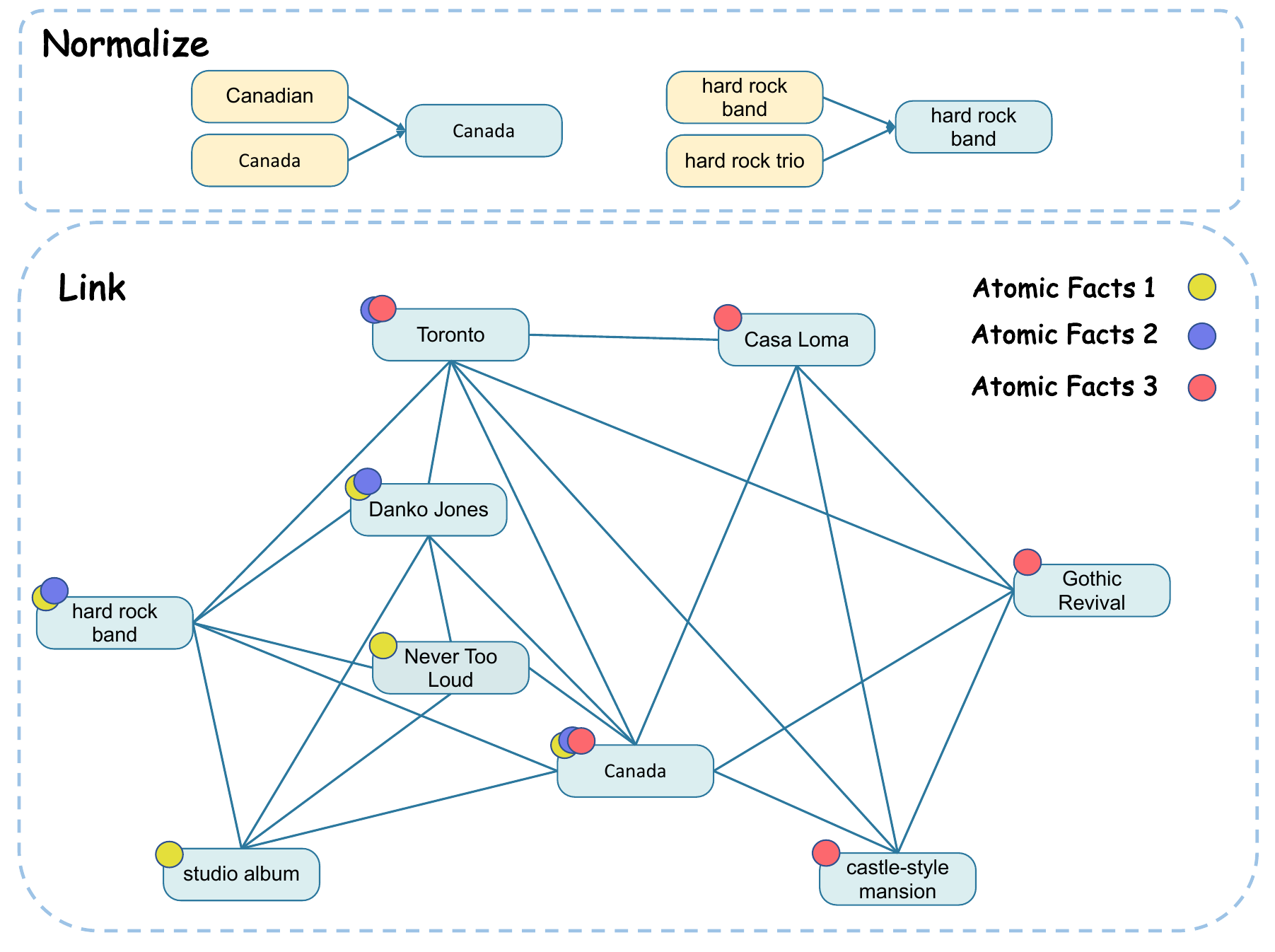}
\end{tcolorbox}

\caption{\textbf{\ourmethod Example(Graph Construction)}. The atomic facts and key elements extracted from the passage correspond to each other, after which the latter are normalized to serve as nodes. Finally, links are formed based on the co-occurrence relationships of the nodes within the atomic facts.}
\label{fig: case_1}
\end{figure*}

\begin{figure*}[!htb]
\begin{tcolorbox}[colback=white,colframe=blue!50!green,title=\textbf{Agent Initialization: Pre-plan And Select Initial Nodes}, center title]

\textbf{Rational Plan}\quad To answer the question, we need to identify the performer or band associated with "Never Too Loud", determine the city where they were formed, and then find out the name of any notable castle in that city.

\textbf{Initial Node}~~~\quad Never Too Loud
\end{tcolorbox}

\begin{tcolorbox}[colback=white,colframe=blue!50!green,title=\textbf{Agent Initialization: Visual Representation}, center title]
\includegraphics[width=\textwidth]{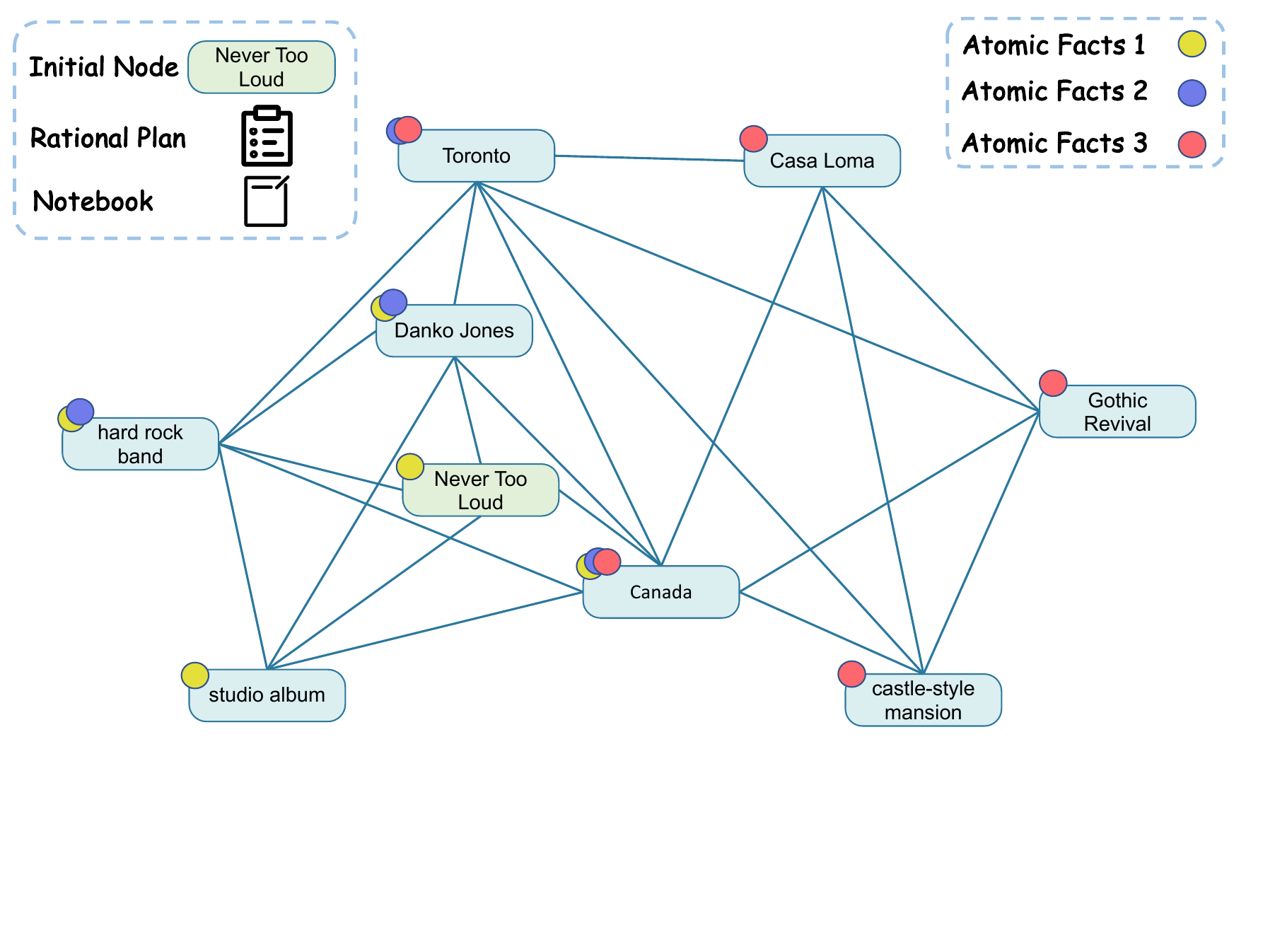}
\end{tcolorbox}

\caption{\textbf{\ourmethod Example(Agent Initialization)}. Initially, a rational plan is formulated in response to the question, guiding further exploration; subsequently, the plan dictates the selection of the initial node from all nodes.}
\label{fig: case_2}
\end{figure*}

\begin{figure*}[!htb]
\begin{tcolorbox}[colback=white,colframe=blue!50!green,title=\textbf{Exploration: Function Call Process}, center title]

\textbf{Exploring Atomic Facts} \quad Node: Never Too Loud; [Atomic Fact 1 from Chunk ID-6]

\textbf{Call Function}\quad \textit{read\_chunk(ID-6)}.

\textbf{Exploring Chunks}\quad  Realized the performer of Never Too Loud is Danko Jones.

\textbf{Call Function:}\quad  \textit{search\_more}

\textbf{Exploring Neighbors}\quad Node: Never Too Loud; Neighbor Nodes: [hard rock band, Danko Jones, studio album, Canada]

\textbf{Call Function}\quad \textit{read\_neighbor\_node(Danko Jones)}

\textbf{Exploring Atomic Facts}\quad Node: Danko Jones; [Atomic Fact 1 from Chunk ID-6, Atomic Fact 2 from Chunk ID-9].

\textbf{Call Function}\quad \textit{read\_chunk(ID-9)}.

\textbf{Exploring Chunks}\quad  Realized Danko Jones band is a band from Toronto, Canada.

\textbf{Call Function:}\quad  \textit{search\_more}

\textbf{Exploring Neighbors}\quad Node: Danko Jones; Neighbor Nodes: [hard rock band, Never Too Lou, studio album, Canada, Toronto]

\textbf{Call Function}\quad \textit{read\_neighbor\_node(Toronto)}

\textbf{Exploring Atomic Facts}\quad Node: Toronto; [Atomic Fact 2 from Chunk ID-9, Atomic Fact 3 from Chunk ID-13].

\textbf{Call Function}\quad \textit{read\_chunk(ID-13)}.

\textbf{Exploring Chunks}\quad  Realized the castle mentioned in the text in Toronto is Casa Loma.

\textbf{Call Function:} \textit{termination}

\end{tcolorbox}

\begin{tcolorbox}[colback=white,colframe=blue!50!green,title=\textbf{Exploration: Visual Representation}, center title]
\includegraphics[width=\textwidth]{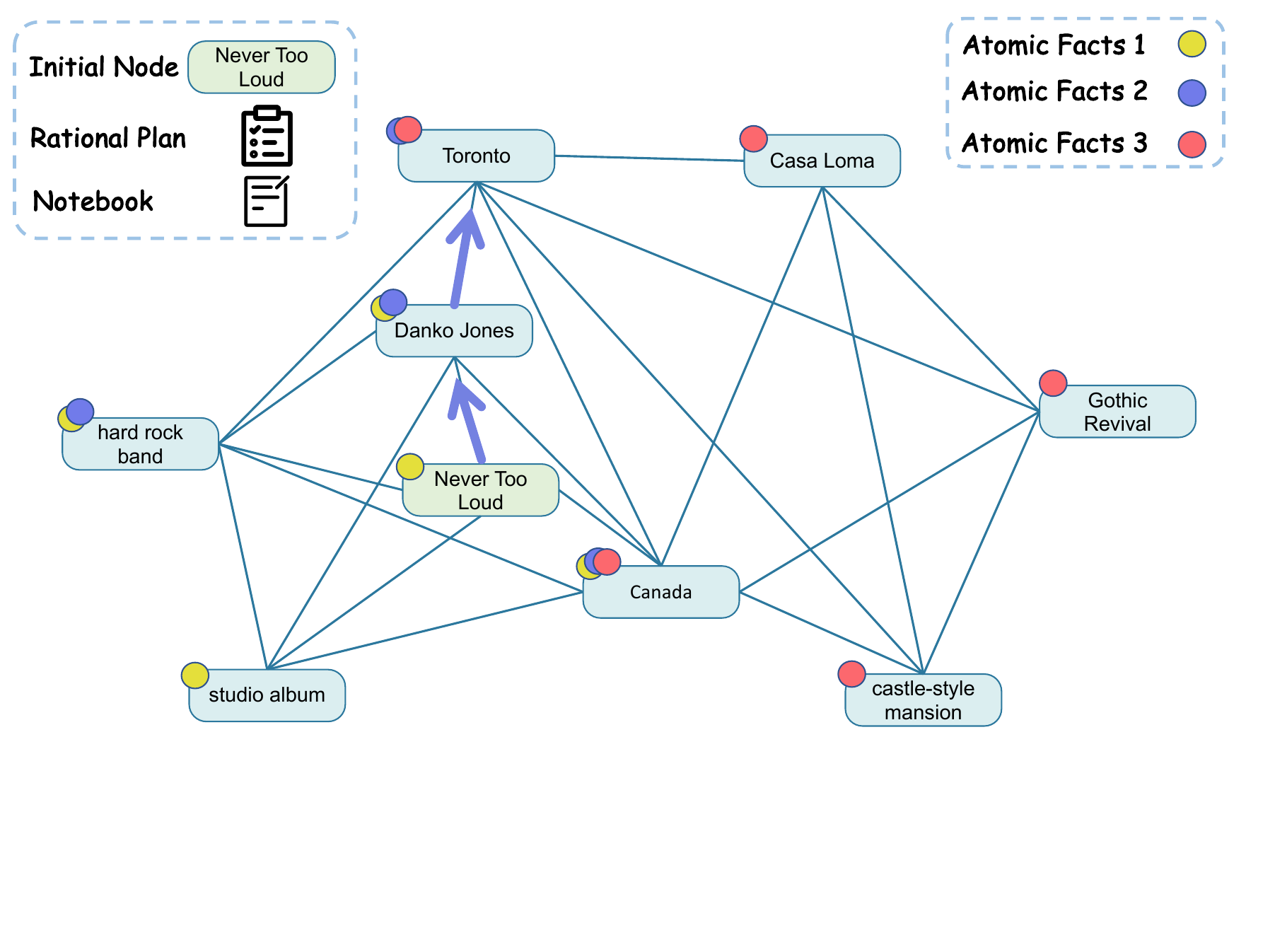}
\end{tcolorbox}

\caption{\textbf{\ourmethod Example(Exploration)}. \ourmethod begins from the initial node, guided by the rational plan, carrying a notebook that records memory, gradually collecting information to answer the question.}
\label{fig: case_3}
\end{figure*}

\vspace{-0.1cm}
This section presents a case study of the \ourmethod workflow. Figure \ref{fig: case_0} displays the posed question alongside the answer and pertinent supporting passages. Subsequently, Figure \ref{fig: case_1} delineates the methodology for constructing the graph. Figure \ref{fig: case_2} further elaborates on the initialization of a pre-planned rational path by \ourmethod and the selection of initial nodes. Figure \ref{fig: case_3} illustrates the sequence of function invocations during the exploration phase. Finally, Figure \ref{fig: case_4} showcases how \ourmethod formulates the answer by leveraging the insights obtained through exploration.
\begin{figure*}[!htb]
\begin{tcolorbox}[colback=white,colframe=blue!50!green,title=\textbf{Question} \& \textbf{Answer}, center title]

\textbf{Question}\quad What is the name of the castle in the city where the performer of Never Too Loud was formed?

\textbf{Answer}~~\quad Casa Loma
\end{tcolorbox}

\begin{tcolorbox}[colback=white,colframe=blue!50!green,title=\textbf{Supporting Passages} ,center title]
\begin{enumerate}
    \item \textbf{Never Too Loud is the fourth studio album by Canadian hard rock band Danko Jones.} It was recorded at Studio 606 in Los Angeles, with the producer Nick Raskulinecz.
    \item \textbf{Danko Jones is a Canadian hard rock trio from Toronto.} The band consists of Danko Jones (vocals/guitar), John "JC" Calabrese (bass), and Rich Knox (drums). The band's music includes elements of hard rock and punk and they are known for their energetic live shows.
    \item \textbf{Casa Loma (improper Spanish for "Hill House") is a Gothic Revival castle-style mansion and garden in midtown Toronto, Ontario, Canada, that is now a historic house museum and landmark.} It was constructed from 1911 to 1914 as a residence for financier Sir Henry Pellatt. The architect was E. J. Lennox, who designed several other city landmarks.
\end{enumerate}

\end{tcolorbox}

\caption{\textbf{\ourmethod Example(Question and Annotations)}. We provide an example question with its answer, along with the supporting passages for this question. This is a typical 3-hop question where we need to gather information and reason step-by-step to arrive at the answer. Details about this example are available at this \href{https://huggingface.co/datasets/THUDM/LongBench/viewer/musique/test?p=1&row=140}{link}.}
\label{fig: case_0}
\end{figure*}

\begin{figure*}[!htb]
\begin{tcolorbox}[colback=white,colframe=blue!50!green,title=\textbf{Answer Reasoning: Response Based on the Notebook}, center title]

\textbf{Question}\quad What is the name of the castle in the city where the performer of Never Too Loud was formed?
\\
\\
\textbf{Memory from the notebook}\quad The performer of Never Too Loud is Danko Jones, which is a band from Toronto, Canada. The text mentions that the castle in Toronto is Casa Loma.
\\
\\
\textbf{\ourmethod answer}~~\quad Casa Loma
\end{tcolorbox}


\caption{\textbf{\ourmethod Example(Answer Reasoning)}. Ultimately, \ourmethod answers the question based on the notebook recorded during the exploration process.}
\label{fig: case_4}
\end{figure*}


\label{sec:appendix}
\end{document}